\documentclass[lettersize,journal]{IEEEtran}
\usepackage{amsmath,amsfonts}
\usepackage{algorithmic}
\usepackage{algorithm}
\usepackage{array}
\usepackage[caption=false,font=normalsize,labelfont=sf,textfont=sf]{subfig}
\usepackage{textcomp}
\usepackage{stfloats}
\usepackage{url}
\usepackage{verbatim}
\usepackage{graphicx}
\usepackage{cite}

\usepackage{orcidlink}
\usepackage{booktabs}
\usepackage{multirow}
\usepackage{makecell}

\begin{document}

\title{XHand: Real-time Expressive Hand Avatar}

\author{Qijun Gan, Zijie Zhou and Jianke Zhu,~\IEEEmembership{Senior Member,~IEEE}
        % <-this % stops a space
\thanks{Qijun Gan, Zijie Zhou and Jianke Zhu are with the College of Computer Science and Technology, Zhejiang University, Zheda Rd 38th, Hangzhou, China. Email: \{ganqijun, zjzhou, jkzhu\}@zju.edu.cn;}% <-this % stops a space
\thanks{Jianke Zhu is the Corresponding Author.}}

\markboth{Journal of \LaTeX\ Class Files,~Vol.~14, No.~8, August~2021}%
{Shell \MakeLowercase{\textit{et al.}}: A Sample Article Using IEEEtran.cls for IEEE Journals}


\maketitle

\begin{abstract}
Hand avatars play a pivotal role in a wide array of digital interfaces, enhancing user immersion and facilitating natural interaction within virtual environments. While previous studies have focused on photo-realistic hand rendering, little attention has been paid to reconstruct the hand geometry with fine details, which is essential to rendering quality. In the realms of extended reality and gaming, on-the-fly rendering becomes imperative. To this end, we introduce an expressive hand avatar, named XHand, that is designed to comprehensively generate hand shape, appearance, and deformations in real-time. To obtain fine-grained hand meshes, we make use of three feature embedding modules to predict hand deformation displacements, albedo, and linear blending skinning weights, respectively. To achieve photo-realistic hand rendering on fine-grained meshes, our method employs a mesh-based neural renderer by leveraging mesh topological consistency and latent codes from embedding modules. During training, a part-aware Laplace smoothing strategy is proposed by incorporating the distinct levels of regularization to effectively maintain the necessary details and eliminate the undesired artifacts. The experimental evaluations on InterHand2.6M and DeepHandMesh datasets demonstrate the efficacy of XHand, which is able to recover high-fidelity geometry and texture for hand animations across diverse poses in real-time. To reproduce our results, we will make the full implementation publicly available at \url{https://github.com/agnJason/XHand}.

% With the advancement of virtual reality, human hands, as crucial interfaces for human-computer interaction, demand meticulous digitization for immersive experiences. While previous works have focused on photo-realistic hand rendering, little attention has been paid to the reconstruction of hand details. In the realms of extended reality and gaming, rendering the fine-grained hand mesh in real-time becomes imperative. To this end, we introduce an expressive hand avatar, named XHand, that is designed to comprehensively generate hand shape, appearance, and deformations in real-time. By making use of three feature embedding modules to predict hand deformation displacements, albedo, and linear blending skinning weights, we obtain fine-grained hand meshes with specific poses. To achieve photo-realistic hand rendering, our method employs a mesh-based neural renderer by leveraging mesh topological consistency and latent codes from embedding modules. During training, a part-aware Laplace smoothing strategy is implemented by incorporating distinct levels of regularization to effectively maintain the necessary details and eliminate undesired artifacts. The experimental evaluations on InterHand2.6M and DeepHandMesh datasets demonstrate the efficacy of XHand, which obtains high-fidelity geometry and texture for hand animations across diverse poses. To reproduce our results, we will make the full implementation publicly available.

\end{abstract}    

\section{Introduction}
\label{sec:intro}

\IEEEPARstart{H}{AND} avatars are crucial in various digital environments, including virtual reality, digital entertainment, and human-computer interaction~\cite{doosti2020hopenet, hasson2020leveraging, fan2021understanding, cheng2015survey}. Accurate representation and lifelike motion of hand avatars are essential to deliver an authentic and engaging user experience. Due to the complexity of hand muscles and the personalized nature, it is challenging to obtain the fine-grained hand representation~\cite{SMPL-X,karunratanakul2022harp,NIMBLE,livehand}, which directly affect the user experience in virtual spaces.

%In human-machine interaction, nimble hands exemplify the potent communication capabilities of humans~\cite{doosti2020hopenet, hasson2020leveraging}. As instruments meticulously designed by nature, the art and science of reconstructing human hands has become a research focus~\cite{SMPL-X,karunratanakul2022harp,NIMBLE,livehand}. Generally, the intricacy of hand muscles and rich details make capturing and modeling hands a challenging endeavor. %As instruments meticulously designed by nature to grasp, gesture, and convey profound meaning, they have garnered unprecedented attention in the realms of science and technology.

Parametric model-based methods~\cite{MANO, SMPL, SMPL-X} have succeeded in modeling digital human, which offer the structured frameworks to efficiently analyze and manipulate the shapes and poses of human bodies and hands. These models have played a crucial role in various applications, enabling computer animation and hand-object interaction~\cite{cao2021reconstructing, MobileHand,alldieck2021imghum, ren2024pyramid, guo20223d}. Since they predominantly rely on mesh-based representations, it restricts them to fixed topology and limited resolution of the 3D mesh. Consequently, it is difficult for these models to accurately represent the intricate details, such as muscle, garments and hair. This hinders them from rendering high fidelity images~\cite{bib:LISA}. 
Model-free methods offer effective solutions for representing hand meshes through various techniques. Graph Convolutional Network (GCN)-based and UV-based representations of 3D hand meshes~\cite{Pose2Mesh, chen2021i2uv} enable the reconstruction of diverse hand poses with detailed deformations. Lightweight auto-encoders~\cite{MobileHand, Moon_2020_ECCV_DeepHandMesh} further enhance real-time hand mesh prediction. Despite these advancements in capturing accurate hand poses, these methods still fall short in preserving intricate geometric details.

Recently, neural implicit representations~\cite{NeRF,wang2021neus} have emerged as powerful tools in synthesizing novel views for static scenes. Some studies~\cite{bib:HumanNeRF, SNARF, liu2021neural, peng2021neural,handnerf, bib:LISA} have expanded these methods into the realm of articulated objects, notably the human body, to facilitate photo-realistic rendering. LiveHand~\cite{livehand} achieves real-time rendering through a neural implicit representation along with a super-resolution renderer. Karunratanakul~\textit{et al.}~\cite{karunratanakul2022harp} present a self-shadowing hand renderer. Corona~\textit{et al.}~\cite{bib:LISA} introduce a neural model LISA that predicts the color and the signed distance with respect to each hand bone independently. Despite the promising results, it struggles to capture intricate high-frequency details and lacks capability of real-time rendering. Meanwhile, Chen~\textit{et al.}~\cite{bib:handavatar} make use of occupancy and illumination fields to obtain hand geometry, while the generated hand geometry lacks the intricate details and appears to be smoothing surface. These methods have difficulties in recovering the detailed geometry that usually plays a crucial role in photo-realistic rendering. 

In addition to hand modeling methods, several studies have focused on reconstructing animatable human bodies or animals~\cite{yang2023reconstructing, luo2022artemis, wu2023magicpony, 10.1145/3528223.3530143,zheng2023pointavatar,zheng2022avatar,grassal2022neural,GaoRecon}. Building accurate human body models presents significant challenges due to the complex deformations involved, particularly in capturing fine details such as textures and scan-like appearances, especially in smaller areas like hands and faces~\cite{SMPL-X, SNARF, yang2022banmo, habermann2021real, peng2021neural, vb-characters}. To address these challenges, several approaches have been developed with detailed 3D scans. For instance, previous works~\cite{bib:HumanNeRF
, peng2022animatable, bhatnagar2020loopreg} have focused on establishing correspondences between pose space and standard space through techniques such as linear blend skinning and inverse skinning weights. These advancements collectively contribute to more precise and realistic human body modeling, while their results for hand modeling remain smooth.

\begin{figure*}[tb]
\hsize=\textwidth
\centering
\includegraphics[width=1.0 \linewidth]{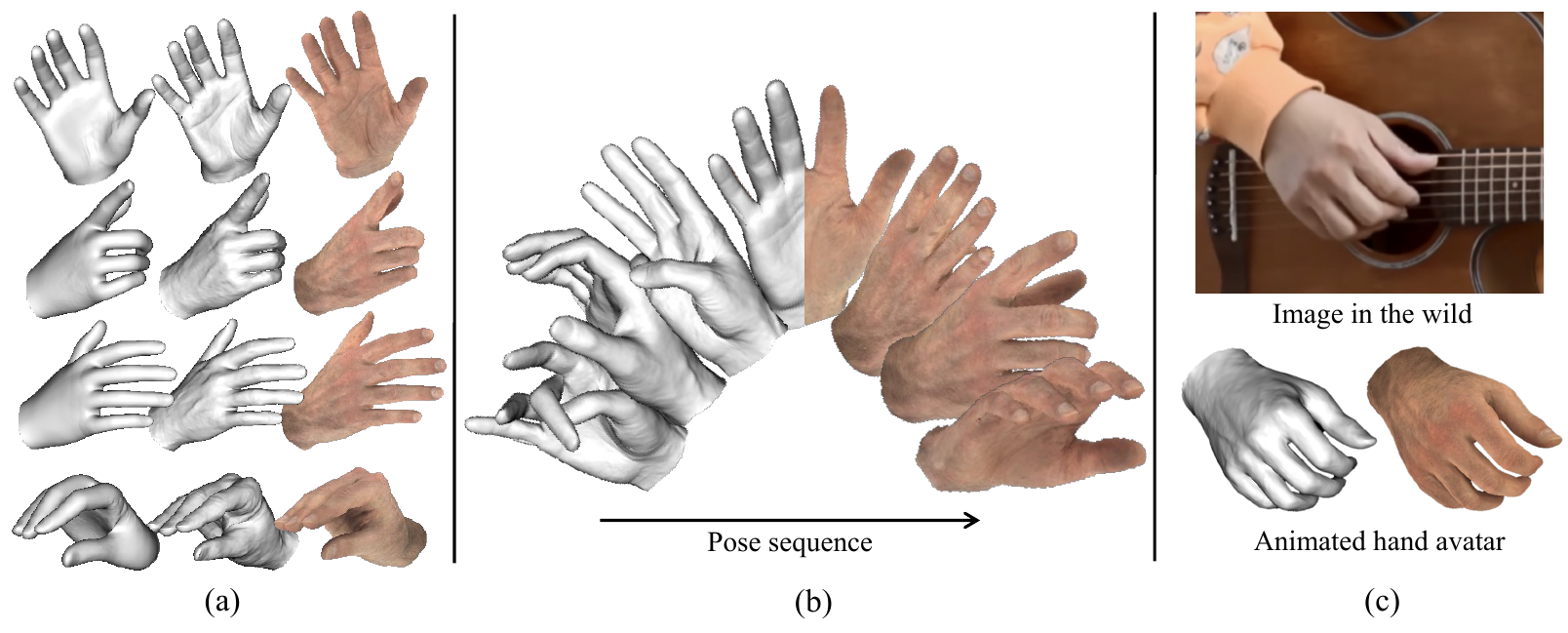}
\caption{We present \textbf{XHand}, a rigged hand avatar that captures the  geometry, appearance and poses of the hand. XHand is created from multi-view videos and utilizes MANO pose parameters (the first image in each group of (a)) to generate high-detail meshes (the second) and renderings (the third). XHand generates photo-realistic hand images in real-time for a given pose sequence (b). (c) is an example of animated personalized hand avatars according to poses~\cite{pavlakos2023reconstructing} in the wild images.}
\label{fig:firstfig}
\end{figure*}

To address these challenges, we propose XHand, an expressive hand avatar that achieves real-time performance (see Fig.~\ref{fig:firstfig}). Our approach includes feature embedding modules that predict hand deformation displacements, vertex albedo, and linear blending skinning (LBS) weights using a subdivided MANO model~\cite{MANO}. These modules utilize average features of the hand mesh and compute feature offsets for different poses, addressing the difficulty in directly learning dynamic personalized hand color and texture due to significant pose-dependent variations. By distinguishing between average and pose-dependent features, our modules simplify the training task and improve result accuracy. Additionally, we incorporate a part-aware Laplace smoothing term to enhance the efficiency of geometric information extraction from images, applying various levels of regularization.

To achieve photo-realistic hand rendering, we use a mesh-based neural renderer that leverages latent codes from the feature embedding modules, maintaining topological consistency. This method preserves detailed features and minimizes artifacts through various regularization levels. We evaluate our approach using the InterHand2.6M dataset~\cite{Moon_2020_ECCV_InterHand2.6M} and the DeepHandMesh collection~\cite{Moon_2020_ECCV_DeepHandMesh}. Experimental results show that XHand outperforms previous methods, providing high-fidelity meshes and real-time rendering of hands in various poses.

Our main contributions are summarized as follows:
\begin{itemize}
     \item A real-time expressive hand avatar with high fidelity results on both rendering and geometry, which is trained with an effective part-aware Laplace smoothing strategy.
     \item An effective feature embedding module to simplify the training objectives and enhance the prediction accuracy by distinguishing invariant average features and pose-dependent features; 
     \item An end-to-end framework to create photo-realistic and fine-grained hand avatars.  The promising results indicate that our method outperforms the previous approaches. 
\end{itemize}

The remainder of this paper is arranged as follows. Related works are introduced in Section~\ref{sec:relate}. The proposed XHand
model and corresponding training process are thoroughly depicted
in Section~\ref{sec:method}. The experimental results and discussion are
presented in Section~\ref{sec:exp}. Finally, Section~\ref{sec:conclusion} sets out the conclusion of this paper
and discusses the limitations.
\section{Relate Work}
\label{sec:relate}

%\textbf{ The related work reads more like a survey of prior work, but not like a critical discussion with respect to this work. How do prior works differ from this one? What are the pros and cons of those with respect to this work? This should all be discussed in the related work section.}

\subsection{Parametric Model-based Method}
3D animatable human models~\cite{SMPL,MANO,SMPL-X} enable shape deformation and animation by decoding the low-dimensional parameters into a high-dimensional space. Loper~\textit{et al.}~\cite{SMPL} introduce a linear model to explicitly represent the human body through adjusting shape and pose parameters. MANO hand model~\cite{MANO} utilizes a rigged hand mesh with fixed-topology that can be easily deformed according to the parameters. The low resolution of the template mesh hinders its application in scenarios requiring higher precision. To address this limitation, Li~\textit{et al.}~\cite{NIMBLE} integrate muscle groups with shape registration, which results in an optimized mesh with finer appearance. Furthermore,
parametric model-based methods~\cite{boukhayma20193d,hasson2019learning,kong2022identityaware,cao2021reconstructing,doosti2020hopenet,hasson2020leveraging, Ren2023EndtoEndWS, sun2023smr,li2021latent} have shown the promising results in accurately recovering hand poses from input images, however, they have difficulty in effectively capturing textures and geometric details for the resulting meshes. In this paper, our proposed XHand approach is able to capture the fine details of both appearance and geometry by taking advantages of Lambertian reflectance model~\cite{DBLP:conf/siggraph/OrenN94}. % Despite this, parametric models remain widely utilized for hand recovery tasks.

\subsection{Model-free Approach}
Parametric models have proven to be valuable in incorporating prior knowledge of pose and shape in hand geometry reconstruction~\cite{MANO}, while their representation capability is restricted due to the low resolution of the template mesh. To address this issue, Choi~\textit{et al.}~\cite{Pose2Mesh} introduce a network based on graph convolutional neural networks (GCN) that directly estimates the 3D coordinates of human mesh from 2D human pose. Chen~\textit{et al.}~\cite{chen2021i2uv} present a UV-based representation of 3D hand mesh to estimate hand vertex positions. Mobrecon~\cite{chen2022mobrecon} predicts hand mesh in real-time through a 2D encoder and a 3D decoder. Despite the encouraging results, the above methods still cannot capture the geometric details of hand. Moon~\textit{et al.}~\cite{Moon_2020_ECCV_DeepHandMesh} propose an encoder-decoder framework that employs a template mesh to learn corrective parameters for pose and appearance. Although having achieved the improved geometry and articulated deformation, it has difficulty in rendering photo-realistic hand images. Gan \textit{et al.}~\cite{gan2024} introduce an optimized pipeline that utilizes multi-view images to reconstruct a static hand mesh. Unfortunately, it overlooks the variations due to joint movements. Karunratanakul~\textit{et al.}~\cite{karunratanakul2022harp} design a shadow-aware differentiable rendering scheme that optimizes the abledo and normal map to represent hand avatar. However, its geometry remains smoothing. In contrast to the above methods, our proposed XHand approach is able to simultaneously synthesize the detailed geometry and photo-realistic images for drivable hands.

%Our work focuses on surpassing the limitation of pose-driven and achieving a drivable hand avatar.

\subsection{Neural Hand Representation}
There are various alternatives available for neural hand representations, such as HandAvatar~\cite{bib:handavatar}, HandNeRF~\cite{handnerf}, LISA~\cite{bib:LISA} and LiveHand~\cite{livehand}. In order to achieve high fidelity rendering of human hands, Chen~\textit{et al.}~\cite{bib:handavatar} propose HandAvatar to generate photo-realistic hand images with arbitrary poses, which take into account both occupancy and illumination fields. LISA~\cite{bib:LISA} is a neural implicit model with hand textures, which focuses on signed distance functions (SDFs) and volumetric rendering. Mundra~\textit{et al.}~\cite{livehand} propose LiveHand that makes use of a low-resolution NeRF representation to describe dynamic hands and  a CNN-based super-resolution module to facilitate high-quality rendering. Despite the efficiency in rendering hand images, it is hard for those approaches to capture the details of hand mesh geometry. Luan~\textit{et al.}~\cite{Luan_2023_CVPR} introduce a frequency decomposition loss to estimate the personalized hand shape from a single image, which effectively address the challenge of data scarcity. Chen~\textit{et al.} introduce a spatially varying linear lighting model as a neural renderer to preserve personalized fidelity and sharp details under natural illumination. Zheng~\textit{et al.} facilitate the creation of detailed hand avatars from a single image by learning and utilizing data-driven hand priors. In this work, our presented XHand method focuses on synthesizing the hand avatars with fine-grained geometry in  real-time.

\subsection{Generic Animatable Objects}
 
In addition to the aforementioned methods on hand modeling, there have been some studies reconstructing animatable whole or partial human bodies or animals~\cite{yang2023reconstructing,luo2022artemis,wu2023magicpony}. Face models primarily pay their attention to facial expressions, appearance, and texture, rather than handling large-scale deformations~\cite{zheng2023pointavatar,zheng2022avatar,grassal2022neural,GaoRecon}. Zheng at al.~\cite{zheng2023pointavatar} bridge the gap between explicit mesh and implicit representations by a deformable point-based model that incorporates intrinsic albedo and normal shading. To build human body model~\cite{SMPL-X,SNARF,zhu2016video,yang2022banmo,habermann2021real,peng2021neural,vb-characters}, numerous challenges arise from the intricate deformations, which make it arduous to precisely capture intricate details, such as textures and scan-like appearances, especially in smaller areas like the hands and face. Previous works~\cite{bib:HumanNeRF,peng2022animatable,bhatnagar2020loopreg} have explored to establish the correspondences between pose space and template space through linear blend skinning and inverse skinning weights. Alldieck~\textit{et al.}~\cite{alldieck2021imghum} employ learning-based implicit representations to model human bodies via SDFs. Chen~\textit{et al.}~\cite{SNARF} propose a forward skinning model that finds all canonical correspondences of deformed points. Shen~\textit{et al.}~\cite{xavatar} introduce XAvatar to achieve high fidelity of rigged human bodies, which employ part-aware sampling and initialization strategies to learn neural shapes and deformation fields. % Our approach introduces feature embeddings to represent hand features and performs a delta shape deformation to a meticulously designed template mesh, employing a forward skinning model as the driving mechanism.
\section{Method}
\label{sec:method}

\begin{figure*}[htp]
    \centering
    \includegraphics[width=1.0 \linewidth]{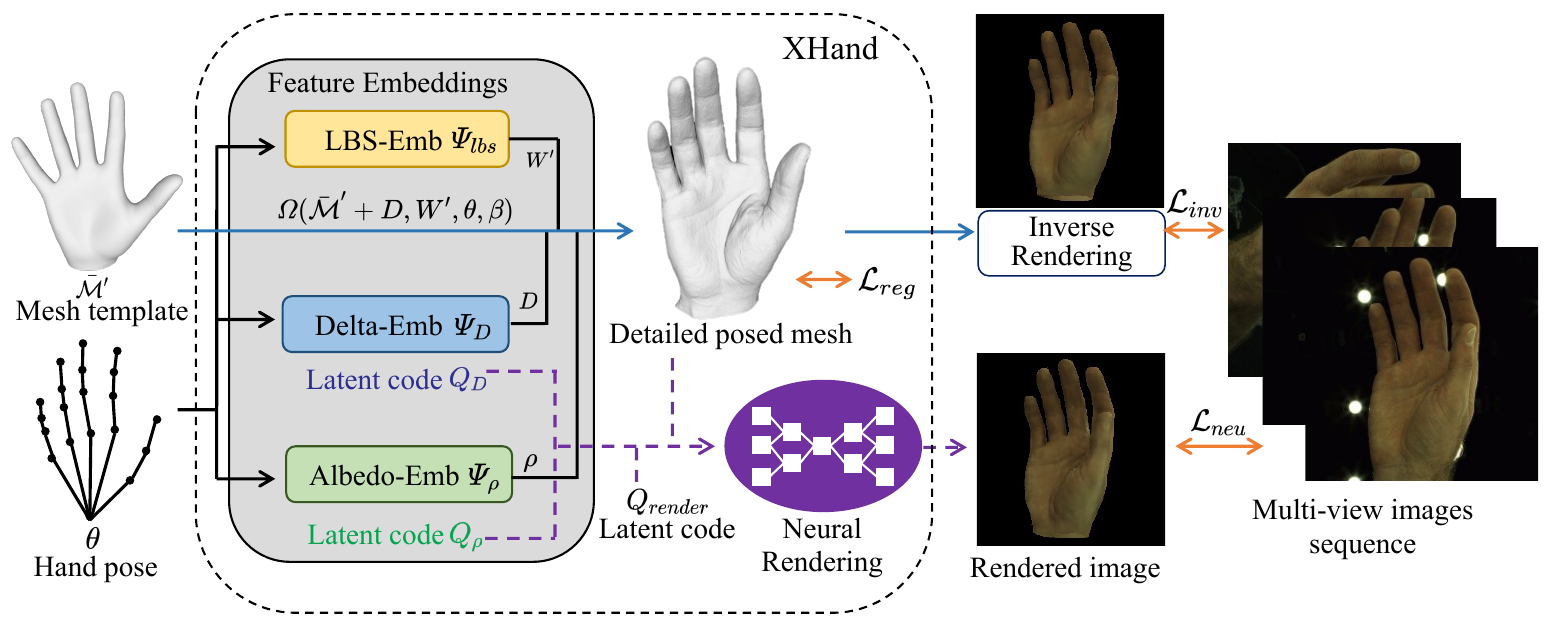}
    \caption{\textbf{Overview of XHand}. Given a hand pose $\theta$, XHand utilizes three feature embedding modules to obtain the displacement field $D$, Linear Blending Skinning (LBS) weights $W$, and albedo $\rho$. These features are applied to the subdivided MANO template $\bar{\mathcal{M}}'$, resulting in a detailed geometric hand mesh. Leveraging mesh-based neural rendering, we achieve photo-realistic renderings. XHand can generate both detailed geometry and realistic rendering in real-time.} 
    \label{fig:overview}
\end{figure*}

Given multi-view images $\{I_{t,i}|i=1,...,N,t=1,...,T\}$ for $T$ frames captured from $N$ viewpoints with pose $\{\theta_t|t = 1,...,T\}$ and shape $\beta$ of their corresponding parametric hand models like MANO~\cite{MANO}, our proposed approach aims to simultaneously recover the expressive personalized hand meshes with fine details and render photo-realistic image in real-time. Fig.~\ref{fig:overview} shows an overview of our method. Given the hand pose parameters $\theta$, the fine-grained posed mesh is obtained from feature embedding modules (Sec.~\ref{sec:xhand}), which are designed to obtain Linear Blending Skinning (LBS) weights, vertex displacements and albedo by combining the average features of the mesh with pose-driven feature mapping. With the refined mesh, the mesh-based neural renderer achieves real-time photo-realistic rendering with respect to the vertex albedo $\rho$, normals $\mathcal{N}$, and latent code $Q$ in feature embedding modules.
%In this section, we initially describe the parametric hand model required to build XHand in Sec.~\ref{sec:mano}, the hand avatar representation in Sec.~\ref{sec:xhand}, and the explanation for how our model can be effectively trained in Sec.~\ref{sec:training}.
%Our approach employs a combination of multiple embeddings. Linear Blending Skinning (LBS) weights are derived from the LBS Embedding $E_{lbs}$, while the Displacement Embedding $E_{D}$ generates displacements for hand skin, additionally, the Albedo Embedding $E_{\rho}$ is utilized to obtain the mesh albedo.

\subsection{Detailed Hand Representation}
\label{sec:xhand}

In this paper, the parametric hand model MANO~\cite{MANO} is employed to initialize the hand geometry, which effectively maps the pose parameter $\theta \in \mathbb{R}^{J\times3}$ with $J$ per-bone parts and the shape parameter $\beta \in \mathbb{R}^{10}$ onto a template mesh $\bar{\mathcal{M}}$ with vertices $V$. Such mapping $\Omega$ is based on linear blending skinning with the weights $W \in \mathbb{R}^{|V|\times J}$. Thus, the posed hand mesh $\mathcal{M}$ can be obtained by 
\begin{equation}
    \mathcal{M} = \Omega(\bar{\mathcal{M}}, W, \theta, \beta).
\end{equation}
%\noindent\textbf{Mesh Subdivision.} To increase mesh resolution and expressive capacity, we perform a uniform subdivision of the MANO template mesh $\bar{\mathcal{M}}$ by introducing new vertices at the midpoints of existing edges. Therefore, we can obtain a high resolution template mesh $\bar{\mathcal{M}}'$ with 49,281 vertices and 98,432 faces. Subsequently, the LBS weights of new vertices are assigned as the average of the seminal vertices.

\begin{figure}[tb]
	\centering
        \includegraphics[width=0.45\textwidth]{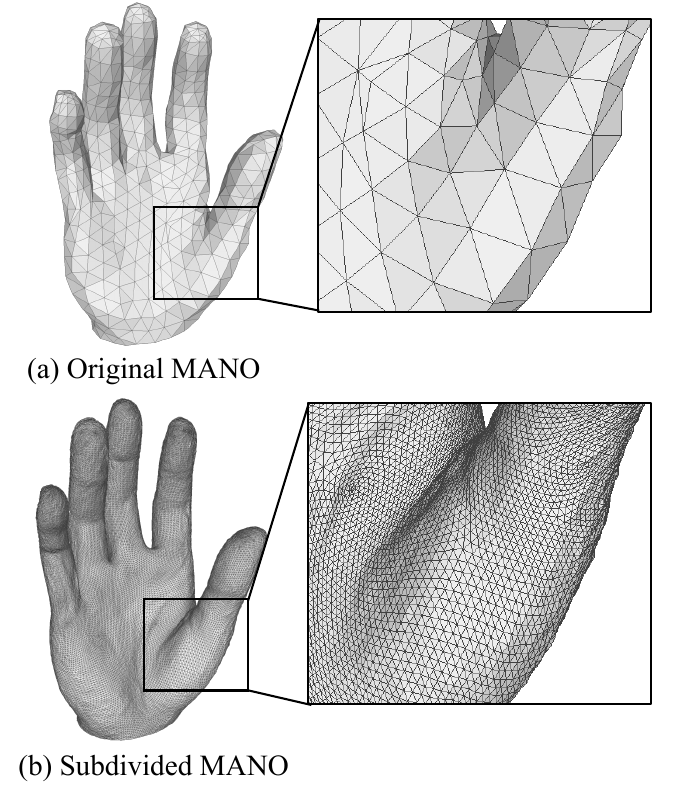}
	\caption{The comparison of mesh refinement and texture between the original MANO and the subdivided MANO.}
	\label{fig:mano}
\end{figure}

\noindent\textbf{Geometry Refinement.} After increasing the MANO mesh resolution for fine geometry using the subdivision method in~\cite{bib:handavatar}, a personalized vertex displacement field $D$ is introduced to allow the extra deformation for each vertex in the template mesh. The refined posed hand mesh $\mathcal{M}_{fine}$ can be computed as below
\begin{equation} \label{eq:mesh}
    \mathcal{M}_{fine} = \Omega(\bar{\mathcal{M}}' + D, W', \theta, \beta).
\end{equation}

The original MANO mesh~\cite{MANO}, consisting of 778 vertices and 1538 faces, has limited capacity to accurately represent fine-grained details~\cite{bib:handavatar}. To overcome this challenge by enhancing the mesh resolution to capture intricate features, we employ an uniform subdivision strategy on the MANO template mesh, as shown in Fig.~\ref{fig:mano}. By adding new vertices at midpoint of each edge for three times, we obtain a refined mesh with 49,281 vertices and 98,432 faces. To associate skinning weights with these additional vertices, we compute the average weights assigned to the endpoints of the corresponding edges.

Let $\mathcal{S}$ denote the subdivision function for MANO mesh. The high resolution template mesh $\bar{\mathcal{M}}'$ and LBS weights $W'$ can be extracted as follows
\begin{equation}
\bar{\mathcal{M}}', W' = \mathcal{S}(\bar{\mathcal{M}}, W).
\end{equation}  

To enhance the fidelity of the hand geometry, the vertex displacements $D$ and the LBS weights $W'$ are pose-dependent for each individual. This enables an accurate representation of the deformation under different poses. To this end, we propose the feature embedding modules $\Psi_{D}$ and $\Psi_{lbs}$ to better capture the intricate details of hand mesh, LBS weights $W'$ are derived from the LBS embedding $\Psi_{lbs}$. The displacement embedding $\Psi_{D}$ generates the vertex displacements $D$. Given the hand pose parameters $\{\theta_{t}|t=1,...,T\}$ for $T$ frames, the mesh features are predicted as follows \begin{equation}
    D_t = \Psi_D(\theta_t), W'_t = \Psi_{lbs}(\theta_t). \label{eq:fe1}
\end{equation}
Thus, the refined mesh $\mathcal{M}_{fine}$ at time $t$ can be formulated as below
\begin{equation}
    \mathcal{M}_{fine} = \Omega(\bar{\mathcal{M}}' + D_t, W'_t, \theta_t, \beta).
\end{equation}

% The three feature embedding modules $E_{lbs}$, $E_{D}$ and $E_{\rho}$ with their corresponding latent code $Q_{lbs}$, $Q_D$ and $Q_\rho$ make use of the same structure of $E(\theta_t | \bar{f}_\mathcal{M})$ except that the target features $f$ are different.

\begin{figure*}[htbp]
	\centering
        \includegraphics[width=1.0\textwidth]{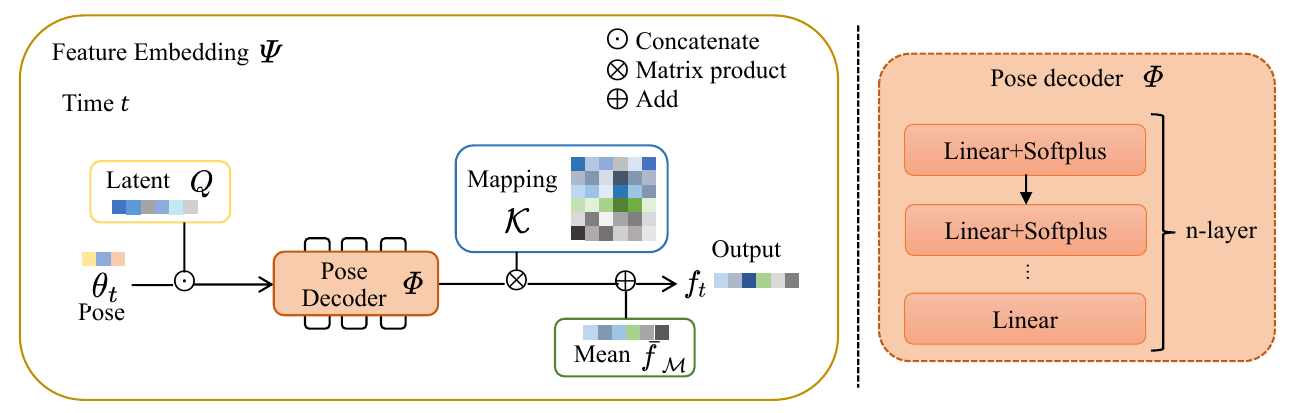}
	\caption{\textbf{Our proposed feature embedding module}. Pose $\theta_t$ at time $t$ is decoded by pose decoder $\Phi$ along with latent code $Q$ for each vertex and mapped to the feature space through mapping matrix $\mathcal{K}$. Finally, it is combined with the average vertex feature $\bar{f}_\mathcal{M}$ to obtain the feature for the pose $\theta$. We introduce three different feature embedding modules $\Psi_{lbs}$, $\Psi_{D}$ and $\Psi_{\rho}$ to predict LBS weights $W'$, displacements $D$ and albedo $\rho$. }
	\label{fig:emb}
    \vspace{-0.10in}
\end{figure*}

%XAvatar~\cite{xavatar} employs a point-sampling approach to train each neural field, which is computationally intensive. In contrast, 
%To simplify the learning objective, we propose an effective feature embedding module to handle various poses. 
\noindent\textbf{Feature Embedding Module.} Generally, it is challenging to learn the distinctive hand features in different poses. To better separate between the deformation caused by changes in posture and the inherent characteristics of the hand, we present an efficient feature embedding module in this paper. It relies on the average features of hand mesh and computes offsets of features in different poses, as illustrated in Fig.~\ref{fig:emb}. 

Given a personalized hand mesh $\mathcal{M}$ and its pose $\theta_t$ at time $t$, our feature embedding module extracts mesh features $f_\mathcal{M}$ as follows \begin{equation}
    f_\mathcal{M} = \Psi(\theta_t | \bar{f}_\mathcal{M}),
\end{equation} 
where $\bar{f}_\mathcal{M}$ denotes the average vertex features of hand mesh. 
%It is challenging for MLPs to discern the influence of specific conditions on mesh features, which may lead to divergence in results. \textbf{WHY?} 

To represent the mesh features of personalized hand generated with hand pose $\theta_t$, we design the following embedding function
\begin{equation}
    \Psi(\theta_t | \bar{f}_\mathcal{M}) = \bar{f}_\mathcal{M} + \Phi(\theta_t, Q) * \mathcal{K},
\end{equation} where $Q$ is vertex latent code to encode different vertices. $\Phi$ denotes a pose decoder that is combined with multi-layer perceptrons (MLPs). It projects the pose $\theta_t$ and latent code $Q$ onto the implicit space. To align with the feature space, $\mathcal{K}$ is the mapping matrix to convert the implicit space $\mathbb{R}^{m}$ into feature space $\mathbb{R}^{n}$, which subjects to \begin{equation}
    \sum_{j=1}^{n} \mathcal{K}_{ij} = 1, \quad \text{for } i = 1, 2, \ldots, m.
\end{equation}

The personalized mesh features $f_\mathcal{M}$ can be derived by combining the average vertex features $\bar{f}_\mathcal{M}$ and the pose-dependent offsets. Consequently, the LBS weights $W_t'$ can be derived with average LBS weights $\bar{f}_{lbs}$, pose decoder $\Phi_{lbs}$, latent code $Q_{lbs}$ and mapping matrix $\mathcal{K}_{lbs}$ as follows
\begin{equation}
    W_t' = \Psi_{lbs}(\theta_t | \bar{f}_{lbs}) = \bar{f}_{lbs} + \Phi_{lbs}(\theta_t, Q_{lbs}) * \mathcal{K}_{lbs}.
\end{equation} Similarly, the vertex displacements $D_t$ can be obtained as follows
\begin{equation}
    D_t = \Psi_{D}(\theta_t | \bar{f}_{D}) = \bar{f}_{D} + \Phi_{D}(\theta_t, Q_{D}) * \mathcal{K}_{D},
\end{equation} where $\bar{f}_{D}$ denotes average displacements. $\Phi_{D}$, $Q_{D}$ and $\mathcal{K}_{D}$ are pose decoder, latent code and mapping matrix for $\Psi_{D}$, respectively. The depths of $\Phi_{lbs}$ within the LBS embedding module $\Psi_{lbs}$ and $\Phi_{\rho}$ within the albedo embedding module $\Psi_{\rho}$ are set to 5, with each layer consisting of 128 neurons. Additionally, the depth of $\Phi_{D}$ within the displacement embedding module $\Psi_{D}$ is 8, where the number of neurons is 512.

{\bf \noindent Remark.} The feature embedding modules allows for the interpretable acquisition of hand features $f_\mathcal{M}$ corresponding to the pose $\theta_t$. The average mesh features are stored in $\bar{f}_\mathcal{M}$, while the features offsets are affected by the pose $\theta$. More importantly, the training objectives are greatly simplified by taking into account of the average features constraints, which leads to the faster convergence and improved accuracy.

\subsection{Mesh Rendering}
\label{render}

\noindent\textbf{Inverse Rendering.} In order to achieve rapid and differentiable rendering of detailed mesh $\mathcal{M}_{fine}$, an inverse renderer is employed to synthesize hand images. Assuming that the skin color follows the Lambertian reflectance model~\cite{sfs}, the rendered image $B$ can be calculated from the Spherical Harmonics coefficients $\mathbf{G}$, the vertex normal $\mathcal{N}$, and the vertex albedo $\rho$ using the following equation \begin{equation}\label{eq:render}
    B(\pi^i) = \rho\cdot SH(\mathbf{G}, \mathcal{N}),
\end{equation} where $\pi^i$ is camera parameter of the $i$-th viewpoint. $SH(\cdot)$ represents Spherical Harmonics (SH) function of the third order. $\mathcal{N}$ is the vertex normal computed from the vertices of mesh $\mathcal{M}_{fine}$. Similar to Eq.~\ref{eq:fe1}, the pose-dependent albedo $\rho_t$ can be obtained from feature embedding module $\Psi_\rho$ with average vertex albedo $\bar{f}_{\rho}$, pose decoder $\Phi_{\rho}$, latent code $Q_{\rho}$ and mapping matrix $\mathcal{K}_{\rho}$ as follows
\begin{equation}
    \rho_t = \Psi_\rho(\theta_t) = \bar{f}_{\rho} + \Phi_{\rho}(\theta_t, Q_{\rho}) * \mathcal{K}_{\rho}.
\end{equation}
By analyzing how the variations in brightness relate to the hand shape, inverse rendering with the Lambertian reflectance model can effectively disentangle geometry and appearance.
%Therefore, we can effectively utilize the geometric and albedo of the mesh to achieve rapid and differentiable rendering.

\noindent\textbf{Mesh-based Neural Rendering.}
The NeRF-based methods usually employ volumetric rendering along its corresponding camera ray $\mathbf{d}$ to acquire pixel color~\cite{handnerf,livehand}, which usually require a large amount of training time. Instead, we aim to minimize the sampling time and enhance the rendering quality by making use of a mesh-based neural rendering method that is able to take advantage of the consistent topology of our refined mesh. % This method leverages the preserved topology of our refined hand representation, resulting in enhanced visual fidelity. 

The mesh is explicitly represented by triangular facets so that the intersection points between rays and meshes are located within the facets. The features that describe meshes, such as position, color, and normal, are associated with their respective vertices. Consequently, the attributes of intersection points can be calculated by interpolating the three vertices of triangular facet to its intersection point. The efficient differentiable rasterization~\cite{Laine2020diffrast} ensures the feasibility of inverse rendering and mesh-based neural rendering.

% as below \begin{equation}
%     \mathcal{C}(\pi^i) = c(\mathbf{x}, \mathcal{N}, \mathbf{h}, \mathbf{d}),
% \end{equation} 

\begin{figure}[tb]
	\centering
        \centering
        % \vspace{-0.1in}
        \includegraphics[width=0.17\textwidth]{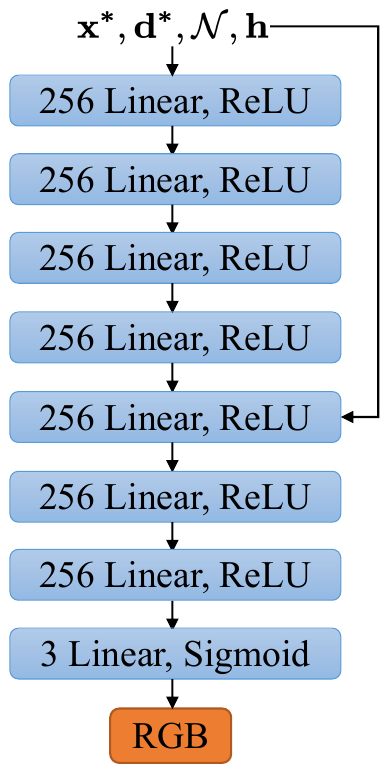}
        % \vspace{-0.1in}
    	\caption{The structure of \textbf{neural renderer} $\mathcal{C}$, where $*$ donates positional encoding~\cite{aliev2020neural}.}
    	\label{fig:nr}
    	\vspace{-0.2in}

\end{figure}

Given a camera view $\pi^i$, our mesh-based neural render $\mathcal{C}(\pi^i)$ synthesizes the image with respect to the position $\mathbf{x}$, normal $\mathcal{N}$, feature vector $\mathbf{h}$ and ray direction $\mathbf{d}$, where $\mathbf{x}$, $\mathbf{h}$ and $\mathcal{N}$ are obtained through interpolating with $\mathcal{M}_{fine}$. $\mathbf{h}$ in neural render $\mathcal{C}(\pi^i)$  contains the latent codes $Q_D$ and $Q_\rho$ detached from $\Psi_D$ and $\Psi_\rho$, and feature vector $Q_{render}$~\cite{gan2024}. $Q_{render}$ is utilized to represent the latent code of vertices during rendering. As in~\cite{NeRF}, the neural network $\mathcal{C}$ comprises 8 fully-connected layers with ReLU activations and 256 channels per layer, excluding the output layer. Furthermore, it includes a skip connection that concatenates the input to the fifth layer, which is depicted in Fig.~\ref{fig:nr}.

% We introduce a rendering network~\cite{chan2022efficient} that can re-render the output through pixel-aligned features in a 3D consistent manner. 

\subsection{Training Process}
\label{sec:training}

To obtain a personalized hand representation, the parameters of the three feature embedding modules $\Psi_D$, $\Psi_{lbs}$, and $\Psi_{\rho}$, as well as the neural render $\mathcal{C}$, require to be optimized based on multi-view image sequences. Our training process consists of three steps, including initialization, training feature embedding modules, and training the mesh-based neural render.

\noindent\textbf{Initialization of XHand.}
To train our proposed XHand model, the average features $\bar{f}_\mathcal{M}$ of mesh in feature embedding significantly affect training efficiency and results. Random initialization has great impact on training due to estimation errors in $\Psi_{lbs}$ and $\Psi_D$, which m ay lead to the failure of inverse rendering. Therefore, it is crucial to initialize the neural hand representation. To this end, the reconstruction result of the first frame ($t=1$) is treated as the initial model. 

Inspired by~\cite{bib:nccsfs,gan2024}, XHand model is initialized from multi-view images. The vertex displacement $D$ and vertex albedo $\rho$ of hand mesh are jointly optimized through inverse rendering. Mesh generation is obtained from Eq.~\ref{eq:mesh}, and the rendering equation is same as Eq.~\ref{eq:render}. The loss function during initialization is formulated as below \begin{equation}\begin{aligned}
    \mathcal{L}_{init} = \sum\limits_{i}||B(\pi^i) - I_i||_1 + \sum L \times D + \sum L \times \rho,
\end{aligned}\end{equation} where $L$ is the Laplacian matrix~\cite{nealen2006laplacian}. Laplacian terms $L \times D$ and $L \times \rho$ are employed to regularize the mesh optimization, as the mesh features are supposed to be smooth. Uniform weights of the Laplacian matrix are adopted in training. The outcomes $D$ and $\rho$ are used to initialize $\Psi_D$ and $\Psi_\rho$. The initialization of $\Psi_{lbs}$ is directly derived from MANO model~\cite{MANO}.

\noindent\textbf{Loss Functions of Feature Embedding.}
Inverse rendering is utilized to learn the parameters of three feature embedding modules $\Psi_D$, $\Psi_{lbs}$ and $\Psi_{\rho}$. $\mathcal{L}_{inv}$ is introduced to minimize the errors of rendering images as follows\begin{equation}
    \mathcal{L}_{inv} = \mathcal{L}_{rgb} + \mathcal{L}_{reg},
\end{equation} where $\mathcal{L}_{rgb}$ represents the rendering loss. $\mathcal{L}_{reg}$ is the regularization term. Inspired by~\cite{kerbl20233d}, we use $L_1$ error combined with an SSIM term to form the $\mathcal{L}_{rgb}$ as below \begin{equation} \begin{aligned} \label{eq:rgb}
    \mathcal{L}_{rgb} &= \lambda\sum\limits_{i}||B(\pi^i) - I_i||_1 + (1-\lambda)\mathcal{L}_{SSIM}(B(\pi^i), I_i),
\end{aligned}\end{equation} where $\lambda$ denotes the trade-off coefficient. %$\mathcal{L}_{inv}$ is used to minimize the loss of inverse rendering, optimizing the parameters of feature embeddings. $\mathcal{L}_{neu}$ minimizes the error of mesh-based neural rendering.%, while the latent code of the feature embeddings is detached, preventing the gradient from propagating forward.

To enhance the efficiency in extracting geometric information from images, we introduce the part-aware Laplace smoothing term $\mathcal{L}_{pLap}$. The Laplace matrix $\mathbf{A}$ of mesh feature $f$ is defined as $\mathbf{A} = L \times f$. Hierarchical weights $\phi_{pLap}$ are introduced to balance the weights of regularisation via different levels of smoothness. $\varphi_{i}$ in matrix $\phi_{pLap}$ is defined as follows \begin{equation}
\varphi_{i} = \left\{\begin{array}{l} \gamma_{1} \quad 0 < \mathbf{A}_{i} < p_1 
\\ \gamma_{2} \quad p_1 < \mathbf{A}_{i} < p_2
\\ ...
\end{array}\right.
,
\end{equation} where $\{p_1, p_2,\ldots \}$ represent the threshold values for the hierarchical weighting and $\{\gamma_1, \gamma_2,\ldots \}$ denote the balanced coefficients. The part-aware Laplace smoothing $\mathcal{L}_{pLap}$ is used to reduce excessive roughness in albedo and displacement without affecting the fine details, which is defined as follows \begin{equation}
    \mathcal{L}_{pLap}(f) = \sum\limits_{i} \phi_{pLap}\mathbf{A}.
\end{equation} By employing varying degrees of hierarchical weights to trade-off Laplacian smoothing, $\mathcal{L}_{pLap}$ is able to better constrain feature optimization in different scenarios. In our cases, minor irregularities are considered to be acceptable, while excessive changes are undesirable. Therefore, the threshold $p$ can be dynamically controlled through the quantiles of Laplace matrix $A$, where those greater than $p$ will be assigned larger balance coefficients.%, such as restricting significant variations while allowing minor changes.

The following regularization terms are introduced to conform the optimized mesh to the hand geometry
\begin{equation} \begin{aligned}
\mathcal{L}_{reg}  = \mathcal{L}_{pLap}(\rho) + \mathcal{L}_{pLap}(D)   + \alpha_{1}\mathcal{L}_{mask} + \alpha_{2}\mathcal{L}_{e} + \alpha_{3}\mathcal{L}_{d}.\label{eq:eq1}
\end{aligned}\end{equation} 
where $\mathcal{L}_{pLap}(\rho)$ and $\mathcal{L}_{pLap}(D)$ are part-aware Laplacian smoothing terms to maintain albedo and displacement flattening during training. $\mathcal{L}_{mask}$, $\mathcal{L}_{e}$ and $\mathcal{L}_{d}$ are utilized to ensure that the optimized hand mesh remains close to the MANO model, where each term is assigned with constant coefficients denoted by $\alpha_1, \alpha_2$ and $\alpha_3$. Let $\mathcal{L}_{mask} = \sum_{i}||\hat{M} - M||_1$ represents the $L_1$ loss between the mask $\hat{M}$ rendered during inverse rendering and the original MANO mask. $\mathcal{L}_{e}$ penalizes the edge length changes of $e_{ij}$  with respect to MANO mesh as $\sum_{i,j}||\hat{e}_{ij} - e_{ij}||_2^2$, where $\hat{e}_{ij}$ is the Euclidean distance $||\cdot||_2^2$ between adjacent vertices $V_i$ and $V_j$ on the mesh edges. $e_{ij}$ denotes the edge distance of the subdivided MANO mesh $\bar{\mathcal{M}}'$. 
$\mathcal{L}_{d} = \sum_{i}||D_i||_2^2$ is employed to constrain the degree of displacement.

\noindent\textbf{Loss Functions of Neural Renderer.}
Once the latent codes $Q_D$ and $Q_\rho$ of $\Psi_D$ and $\Psi_\rho$ are detached, $\mathcal{L}_{neu}$ is used to minimize the residuals between the rendered image and the ground truth like Eq.~\ref{eq:rgb}\begin{equation} %\begin{aligned}
    \mathcal{L}_{neu} = \omega\sum\limits_{i}||\mathcal{C}(\pi^i) - I_i||_1 + (1-\omega)\mathcal{L}_{SSIM}(\mathcal{C}(\pi^i), I_i),
%\end{aligned}
\end{equation}
where $\omega$ denotes balanced coefficient.
\section{Experiments}
\label{sec:exp}
%\textbf{geometry evaluation; compare with HARP}

%In this section, we introduce the implementation details and evaluation metrics in Sec.~\ref{subsec:details}, and compare the geometric and rendering approaches with state-of-the-art methods in Sec.~\ref{subsec:exp}. Finally, in Sec.~\ref{subsec:abl}, we conduct ablative experiments to demonstrate the effectiveness of the modules we designed.

\subsection{Datasets}

\noindent\textbf{InterHand2.6M.} The InterHand2.6M dataset~\cite{Moon_2020_ECCV_InterHand2.6M} is a large collection of images, each with a resolution of $512\times334$ pixels, accompanied by MANO annotations. It includes multi-view temporal sequences of both single and interacting hands. The experiments primarily utilize the 5 FPS version of this dataset.

\noindent\textbf{DeepHandMesh.} The DeepHandMesh dataset~\cite{Moon_2020_ECCV_DeepHandMesh} features images captured from five different viewpoints, matching the resolution of those in InterHand2.6M. It also provides corresponding 3D hand scans, facilitating the validation of mesh reconstruction quality against 3D ground truth data.

\subsection{Experimental Setup}
\label{subsec:details}
\noindent\textbf{Implementation Details.} In the experiments, our proposed XHand model is mainly trained and evaluated on the 5FPS version of Interhand2.6M dataset~\cite{Moon_2020_ECCV_InterHand2.6M}, which is made of large-scale multi-view sequences capturing a wide range of hand poses. Each sequence has dozens of images with the size of $512\times334$. As in~\cite{bib:handavatar, handnerf}, XHand model is trained on the InterHand2.6M dataset with 20 views across 50 frames for each sequence. The remaining frames are used for evaluation. To assess the quality of mesh reconstruction, we conduct experiments on DeepHandMesh dataset~\cite{Moon_2020_ECCV_DeepHandMesh}, which consists of 3D hand scans along with images captured from five different views. The images are with the same size as those in InterHand2.6M dataset. We conducted all the experiments on a PC with NVIDIA RTX 3090 GPU having 24GB GPU memory.

%The input of neural renderer consists four components, including the vertex position $\mathbf{x}$, the normal vector $\mathbf{n}$, the feature vector $\mathbf{h}$, and the ray direction $\mathbf{d}$.
\begin{figure*}[tb]
    \centering
    \includegraphics[width=1.0 \linewidth]{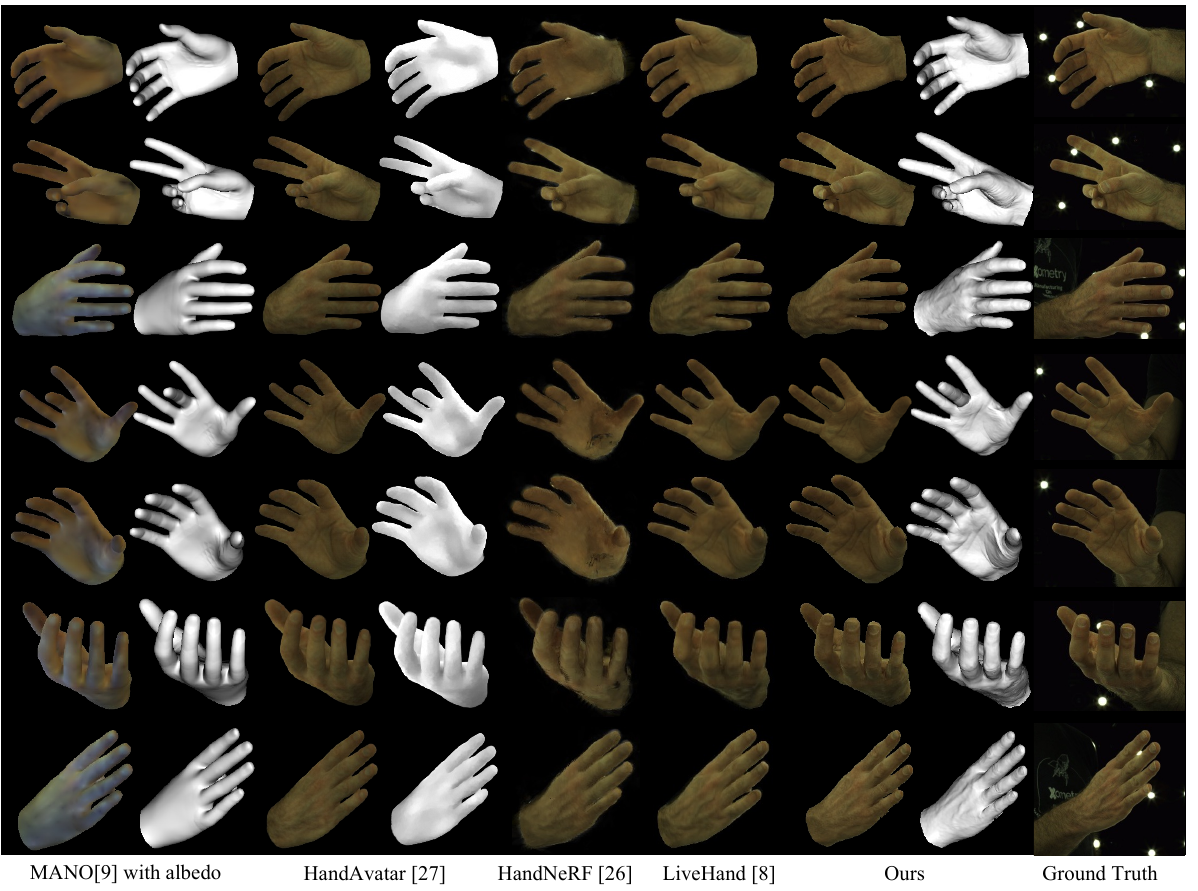}
    % \vspace{-0.25in}
    \caption{\textbf{Visual results on image synthesis.} We show the rendering results of single hand, which are optimized and trained from InterHand2.6M~\cite{Moon_2020_ECCV_InterHand2.6M}. The hands rendered with pure white color represent the shading in order to highlight the level of mesh detail. The visualizations of HandNeRF~\cite{handnerf} are provided by its authors.}
    \label{fig:vis}
    \vspace{-0.1in}
\end{figure*}

We employ PyTorch and Adam Optimizer with a learning rate of $5e^{-4}$. To facilitate differentiable rasterization, we make use of the off-the-shelf renderer nvdiffrast~\cite{Laine2020diffrast}. As in~\cite{aliev2020neural}, positional encoding is performed on $\mathbf{d}$ and $\mathbf{x}$ before feeding them into the rendering network. In our training process, the feature embedding modules are firstly trained for 500 epochs using inverse rendering. Then, feature embedding modules and neural render are jointly trained for 500 epochs, where the average features $\bar{f}_\mathcal{M}$ in feature embedding modules are updated every 50 epochs. We empirically found that the best performance is achieved in case of $\lambda=\omega=0.8$, $\alpha_1=10$, $\alpha_2=1e^5$, and $\alpha_3=1e^4$. To avoid the excessive displacements and color variations, in $\mathcal{L}_{pLap}(\rho)$, $p_1$ is set to the first quartile of $A_\rho$, $\gamma_{1}$ is set to $0.1$, and $\gamma_{2}$ is $1$. Similarly, in $\mathcal{L}_{pLap}(D)$, $p_1$ is the median of $A_D$, and $\gamma_{1} = 0.1$, $\gamma_{2} = 20$. The lengths of latent codes $Q_{lbs}$, $Q_D$, $Q_\rho$ and $Q_{render}$ are set to 10, 10, 10 and 20, respectively.

\noindent\textbf{Evaluation Metrics.} In the experiments, we fit the hand mesh representations to multi-view images sequence for single scene. For fair comparison, we employ the same evaluation metrics as in~\cite{livehand, bib:handavatar, handnerf}, which measure the synthesized results with peak signal-to-noise ratio (PSNR), structural similarity index (SSIM), and learned perceptual image patch similarity (LPIPS). We calculate the average point-to-surface Euclidean distance (P2S) to assess the accuracy of the reconstructed hand mesh, which is measured in millimeters since the Chamfer distance metric is considered unsuitable due to scale variations between MANO and 3D scans.

\subsection{Experimental Results}
\label{subsec:exp}

To investigate the efficacy of our proposed XHand, we treat the subdivided MANO model~\cite{MANO} with vertex albedo as our baseline, which has the merits of the efficient explicit representation. Moreover, we compare our model against several rigged hand expression methods, including LISA~\cite{bib:LISA}, HandAvatar~\cite{bib:handavatar}, HandNeRF~\cite{handnerf}, and LiveHand~\cite{livehand}. For fair comparison, LiveHand is re-trained with the same setting and LISA is reproduced by~\cite{livehand}.% with fair comparison. 

\begin{table}[tb]
    \centering
    \caption{Rendering quality comparisons on the InterHand2.6M dataset. Our method excels in delivering the best rendering quality while simultaneously maintaining real-time performance.}
    \renewcommand\arraystretch{1.2}
	\begin{tabular}{c | c c c c}
            \Xhline{1pt}
            Model & LPIPS $\downarrow$ & PSNR $\uparrow$ & SSIM $\uparrow$ & FPS $\uparrow$\\
            \hline
            MANO~\cite{MANO} with abledo & 0.026 & 28.56 &  0.972 & \bf306.0 \\
		  HandAvatar~\cite{bib:handavatar}&  0.050 & 33.01   & 0.933 & 0.2 \\
    	  LISA~\cite{bib:LISA} &  0.078 & 29.36 & - & 3.7 \\
            HandNeRF~\cite{handnerf}  &  0.048 & 33.02  & 0.974 & - \\
            LiveHand~\cite{livehand}  &  0.025 & 33.79  & 0.985 & 45.5 \\
            Ours  & \bf0.012  & \bf34.32  & \bf0.986 & 56.2 \\
            \Xhline{1pt}
	\end{tabular}
    %\vspace{-0.1in}
 \label{tab:compare}
\end{table}

\begin{table}[tb]
\centering
    \caption{Evaluation of mesh reconstruction quality on DeepHandMesh dataset with 5 views. We report the mean P2S(mm) of each sequence.}
    \renewcommand\arraystretch{1.2}
    \setlength{\tabcolsep}{2mm}{
	\begin{tabular}{c| c c c | c}
            \Xhline{1pt}
                Model & Rigid fist & Relaxed  & Thumb up & Average \\
            \hline 
            MANO~\cite{MANO} &  6.469 & 5.719 & 5.224 & 5.659 \\
		  DHM~\cite{Moon_2020_ECCV_DeepHandMesh} & 2.695 & 3.995 & 3.639 & 3.492 \\
	      Ours &  \bf2.593 & \bf2.189 & \bf2.162 & \bf2.276 \\
            \Xhline{1pt}
	\end{tabular}}
\label{tab:mesh}
%\vspace{-0.10in}
\end{table}

We firstly perform the quantitative evaluation on rendering quality, as shown in Table~\ref{tab:compare}. The evaluation metrics of LISA~\cite{bib:LISA} are adopted from LiveHand~\cite{livehand} and the results of HandNeRF~\cite{handnerf} are obtained from their original paper. It can be seen that our proposed XHand approach achieves the best results with a PSNR of 34.3dB. Our baseline drives a textured MANO model through LBS weights. Due to lacking the ability to handle illumination changes across different scenes and poses, there exist some artifacts with a PSNR of 28.6dB. NeRF-based methods~\cite{bib:LISA, bib:handavatar, handnerf, livehand} present the competitive PSNR results, which rely on MANO mesh without fine-grained geometry during rendering. By taking advantage of fine-grained meshes estimated by XHand, our method outperforms the previous approaches using volumetric representation in terms of the rendering quality. Benefiting from our design, XHand achieves 56 frames per second (FPS) on inference. Specifically, the feature embedding modules require 0.7 milliseconds, inverse rendering requires 15 milliseconds and the neural rendering module needs 0.1 milliseconds.%LISA~\cite{bib:LISA} employs a radiance field to learn hand appearance while having difficulties in accurately capturing hand pose with a learnable skinning-based deformation. This is because XHand takes advantage of the topological consistent mesh model to generate stable and high fidelity images.  and outperforms the state-of-the-art methods at large margin.

%Benefiting from our design, our method is able to drive high-fidelity geometry in real-time and generate hyper-realistic images .  In terms of image rendering, by taking advantage of the topological consistency, our method can generate stable and robust rendered images, as shown in Table~\ref{tab:compare}. In addition, our method achieves a speed of 46 FPS for inferring a single frame image. The feature embedding modules require 14.3 milliseconds, while the neural rendering module averages at 4.1 milliseconds.

\begin{figure}
	\centering
        \includegraphics[width=0.9 \linewidth]{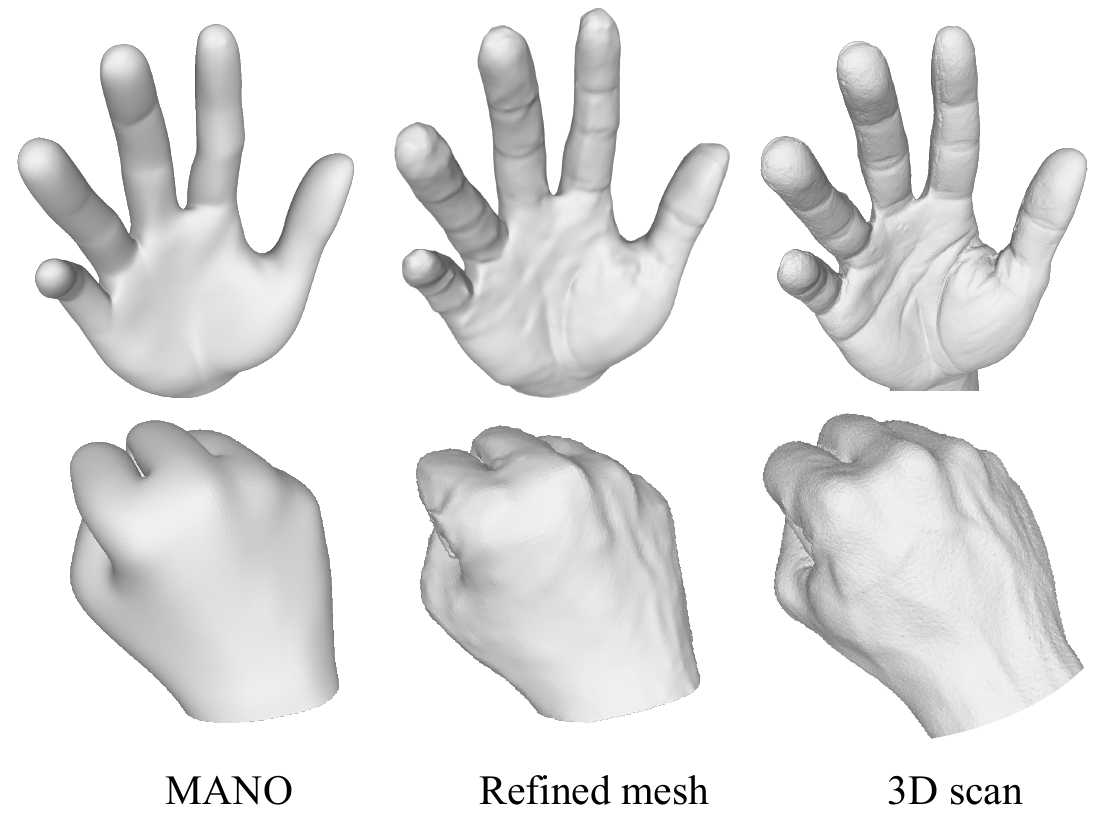}
	\caption{More visual results on DeepHandMesh~\cite{Moon_2020_ECCV_DeepHandMesh}. The proposed method produces highly detailed hand models that capture intricate features such as folds and textures.}
	\label{fig:dhm}
\end{figure}

Table~\ref{tab:mesh} shows the results on DeepHandMesh dataset. Our method outperforms the annotated MANO mesh~\cite{MANO} and DHM~\cite{Moon_2020_ECCV_DeepHandMesh} by 3.3 mm and 1.2 mm on P2S. This indicates that our proposed feature embedding module can accurately capture the underlying hand mesh deformation comparing to the encoder-decoder scheme in DHM.
More experimental results conduct on the DeepHandMesh~\cite{Moon_2020_ECCV_DeepHandMesh} dataset are visualized in Fig.~\ref{fig:dhm}. % TODO!! refine and refig

\begin{figure}[tb]
    \centering
    \includegraphics[width=1.0 \linewidth]{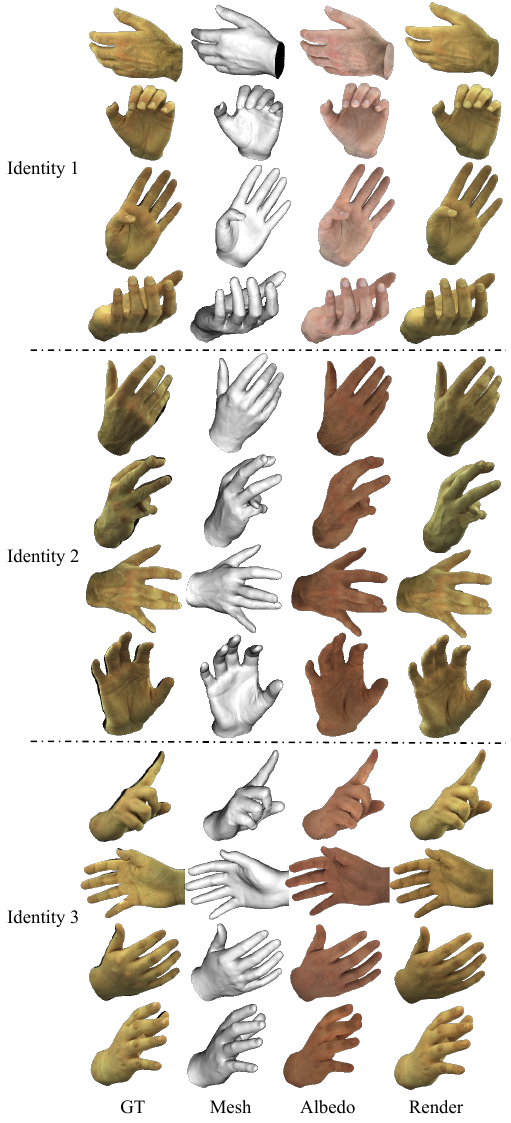}
    % \vspace{-0.1in}
    \caption{\textbf{Visual results of different identities}. XHand has the capability to drive the hand avatar of different persons through different poses.}
    \label{fig:idt}
    % \vspace{-0.2in}
\end{figure}

For better illustration, Fig.~\ref{fig:vis} shows the more detailed comparisons of rendering and geometry on InterHand2.6M test split. Due to the limited expressive capability, it is hard for the baseline MANO model~\cite{MANO} to capture muscle details varying across different poses. Although the hand meshes generated by HandAvatar~\cite{bib:handavatar} have more details than MANO, they are still smoothing compared to ours. In terms of geometry, our method exhibits more prominent skin wrinkles based on different poses. The NeRF-based method HandNeRF~\cite{handnerf} and LiveHand~\cite{livehand} yield the competitive render results, while they still rely on the MANO model and cannot obtain fine-grained hand geometry. On the contrary, our approach effectively presents an accurate hand representation by taking advantage of the feature embedding module and the topological consistent mesh model, resulting in enhanced rendering quality and geometry quality. Fig.~\ref{fig:idt} visualizes the results of different identities animated using reference poses. 

 % TODO!! maybe change?

\begin{figure*}[htp]
	\centering
         \vspace{-0.1in}
        \includegraphics[width=1.0 \linewidth]{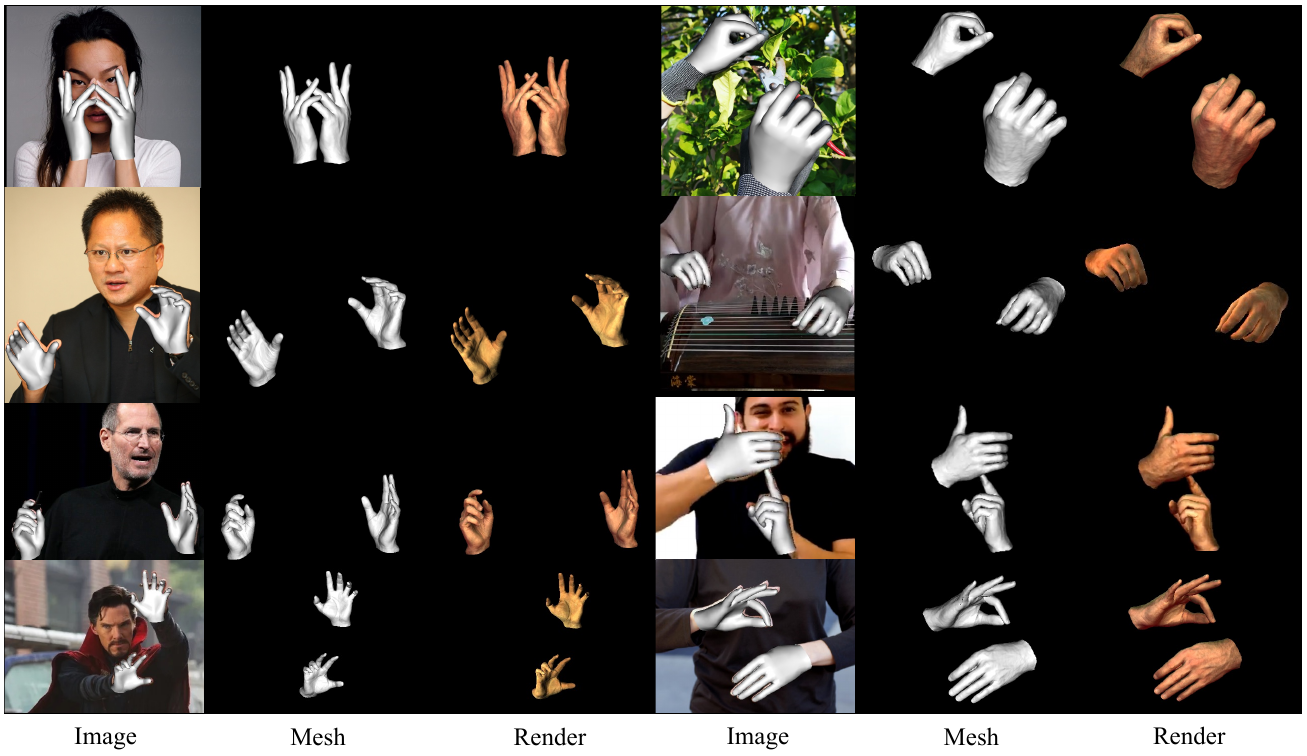}
        \vspace{-0.05in}
	\caption{More visual results on wild images. The MANO parameters are extracted from Hamer~\cite{pavlakos2023reconstructing}. The proposed method is capable of generating personalized hand expressions based on any given hand gesture. This approach allows for the accurate and efficient translation of a wide range of hand poses into detailed, individualized hand representations, ensuring high fidelity and adaptability across various input gestures. }% 
	\label{fig:wild}
  \vspace{-0.1in}
\end{figure*}

The proposed method efficiently drives personalized hand expressions from arbitrary hand gesture inputs. To demonstrate its performance, in-the-wild data serve as a reference for hand poses, as illustrated in Fig.~\ref{fig:wild}. The pose parameters of in-the-wild videos are extracted from HaMeR~\cite{pavlakos2023reconstructing}.
It is worth noting that we can enhance the vividness of the images by using different spherical harmonic coefficients for relighting.  % TODO!! add wild data

\subsection{Ablation Study}
\label{subsec:abl}

\begin{figure*}[htb]
    \centering
    \includegraphics[width=1.0 \linewidth]{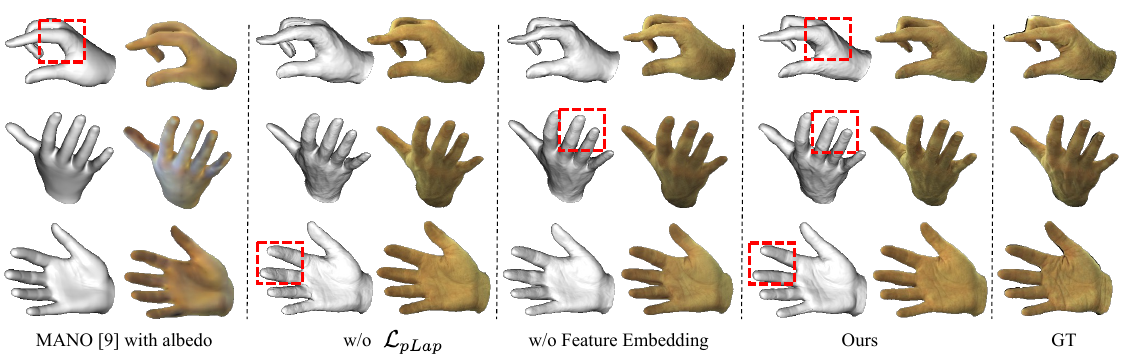}
    \vspace{-0.1in}
    \caption{\textbf{Ablation visual results on the test set}. We present visual results across different modules. Our method yields highly detailed results (see red regions). Note that our results are more realistic by implicitly controlling deformations with respect to hand poses.} % The regularization term $\mathcal{L}_{pLap}$ allows us to remove artifacts without sacrificing details. Feature embedding modules provide us with enhanced level of detail.
    \label{fig:abl}
     \vspace{-0.1in}
\end{figure*}

We perform extensive ablation experiments on the InterHand2.6M dataset test set to validate the contributions of various modules and settings within our framework. First, we aim to demonstrate the performance improvements achieved by our proposed feature embedding module and part-aware Laplace smoothing strategy, consistent with our design intentions for the fusion modules. Second, we intend to showcase the robust performance of our XHand model across different numbers of views, highlighting its effectiveness even with limited viewpoints. Furthermore, we conduct a comparative analysis of various neural rendering networks. Based on this evaluation, we have chosen MLPs to enhance both the inference speed and the rendering quality, ensuring efficient and high-fidelity output. The following sections detail these ablation experiments and analyze the results comprehensively.

\noindent\textbf{Ablation Study on Different Components.} In the first row of Fig.~\ref{fig:abl}, it can be seen that our method significantly highlights skeletal movements and skin changes. Moreover, our design resolves the issue of lighting variations. Our proposed part-aware Laplacian regularization effectively reduces the surface artifacts without sacrificing the details. The feature embedding modules are able to guide the learning of hand avatars by distinguishing average features and pose features, which enhance the reconstruction accuracy.

\begin{table}[tb]
	\centering
		\centering
        \caption{Ablation study of our XHand on InterHand2.6M~\cite{Moon_2020_ECCV_InterHand2.6M}. The effects of different components are evaluated.}
        \renewcommand\arraystretch{1.2}
        {
        \begin{tabular}{c | c c c}
            \Xhline{1pt}
            Model & LPIPS $\downarrow$ & PSNR $\uparrow$ & SSIM $\uparrow$\\
            \hline 
            MANO~\cite{MANO} with abledo & 0.0257 & 28.56 &  0.9715  \\
		  w/o feature embedding&  0.0139 & 32.81   & 0.9838   \\
    	  w/o $\mathcal{L}_{pLap}$ & 0.0129  & 32.87 & 0.9843 \\
            w/o Position Encoder  & \bf0.0114 & 33.95  & 0.9853 \\
             \textbf{Ours}  & 0.0123  & \bf34.32  & \bf0.9859  \\
            \Xhline{1pt}
	   \end{tabular}}
		\label{tab:abl}
\end{table}
\begin{table}[tb]
		\centering
        \caption{Ablation study on different number of views. } %More viewpoints lead to better results. Even with a small number of viewpoints, our method ensures a certain level of rendering capability.
        \renewcommand\arraystretch{1.2}	
         \begin{tabular}{c | c c c}
            \Xhline{1pt}
            Num views & LPIPS $\downarrow$ & PSNR $\uparrow$ & SSIM $\uparrow$\\
            \hline 
            1-view & 0.0209 & 29.34 &  0.9712  \\
		  5-view & 0.0135  &  32.72  &  0.9823  \\
    	  10-view &  0.0129 & 33.50 & 0.9832 \\
            20-view  & 0.0123  & 34.32  & 0.9859 \\
            30-view  & \bf0.0091  & \bf35.23  & \bf0.9865 \\
            \Xhline{1pt}
	\end{tabular}
		\label{tab:views}
 \vspace{-0.1in}
\end{table}

Table~\ref{tab:abl} shows that the level of mesh detail significantly affects image quality. The rendering results are substantially enhanced through feature embedding. The part-aware Laplacian regularization yields more realistic geometric results, indirectly improving the accuracy of the neural render. Furthermore, the Position Encoder in neural rendering leads to better image quality. %From the second row of Fig.~\ref{fig:abl}, it can be observed that this design brings a fine mesh.

\noindent\textbf{Ablation Study on Number of Views.} Typically, the performance of each model is improved along with the increasing number of input images, particularly for the NeRF-based methods. Also, insufficient training data may lead to the reconstruction failure. We conducted ablation experiments using different numbers of views as inputs. As shown in Table~\ref{tab:views}, we trained the model on sequences of 1, 5, 10, 20 and 30 views to demonstrate the impact of views. Despite being trained with a limited number of viewpoints, including as few as a single viewpoint, our method effectively captures the hand articulations. Furthermore, we achieve the competitive results in case of more than 10 input views.

\begin{table}[tb]
    \centering
    \vspace{-0.1in}
    \caption{Rendering quality and inference speed comparisons among MLPs, UNet and EG3D~\cite{chan2022efficient} used in neural rendering.}
    \renewcommand\arraystretch{1.2}
	\begin{tabular}{c | c c c c}
            \Xhline{1pt}
            \textbf{Method} & LPIPS $\downarrow$ & PSNR $\uparrow$ & SSIM $\uparrow$ & FPS $\uparrow$ \\
            \hline 
            XHand-MLPs  & 0.012  & 34.32  & 0.986 & \bf56.2 \\
            XHand-UNet  & \bf0.011  & \bf34.72  & \bf0.987 & 46.2 \\
            XHand-EG3D~\cite{chan2022efficient}  & 0.013  & 32.3  & 0.981 & 40.4 \\
            \Xhline{1pt}
	\end{tabular}
  \vspace{-0.1in}
 \label{tab:mlp-unet}
\end{table}

\noindent\textbf{Choices of Neural Rendering.} Traditional neural radiance fields~\cite{NeRF} typically employ 8-layer MLPs as the renderer. In contrast, our mesh-based network eliminates the necessity for point cloud sampling, which is able to render through vertex features. Benefiting from topology consistency, our neural renderer can make use of UNet~\cite{unet} which leads to promising performance. To explore this, we conduct ablation experiments on both network architectures, as detailed in Table~\ref{tab:mlp-unet}. These experimental results demonstrate that a UNet with 4 layers achieves superior rendering quality, albeit at the expense of inference speed. In comparison to UNet, MLPs can enhance performance by 20\% with only a marginal loss in accuracy. Therefore, we have chosen to employ MLPs as our neural renderer. Furthermore, our investigation into a well-designed image generation network, EG3D~\cite{chan2022efficient}, reveals its unsuitability for neural rendering. % TODO!! read and refine

\section{Conclusion}
\label{sec:conclusion}
We present XHand, a real-time expressive hand avatar with photo-realistic rendering and fine-grained geometry. By taking advantage of the effective feature embedding modules to distinguish average features and pose-dependent features, we obtain the finely detailed meshes with respect to hand poses. To ensure the high quality of hand synthesis, our method employs a mesh-based neural render that takes consideration of mesh topological consistency. During the training process, we introduce the part-aware Laplace regularization to reduce the artifacts while maintaining the details through different levels of regularization. Rigorous evaluations conducted on the InterHand2.6M and DeepHandMesh datasets demonstrate the ability to produce high-fidelity geometry and texture for hand animations across a wide range of poses.

%\noindent\textbf{Limitations and future work.} 
Our method relies on accurate MANO annotations provided by the dataset during training. For future work, we will consider to explore the effective MANO model parameter estimator. %Furthermore, our method is based on the assumption of diffuse reflection to learn skin details, which does not perform well in the presence of strong specular reflections. We will explore surface reconstruction for different materials to address this limitation. 

%This indicates that the unlabeled datasets are hard to train. Existing solutions for pose estimation have limited accuracy in different scenes, highlighting the value of ongoing research on pose estimation from image sequences.%Pose estimation from view sequencesthe is worthy of ongoing exploration.% Furthermore, our method is based on the assumption of diffuse reflection to learn skin details, which does not perform well in the presence of strong specular reflections. We will continue to explore surface reconstruction for different materials to address this limitation. 

%Our method relies on the MANO pose provided by the dataset, which indicates that other unlabeled datasets~\cite{Moon_2020_ECCV_DeepHandMesh} are not able to be used for training XHand. Additionally, the accuracy of the provided pose has a significant impact on the training results. Existing solutions for estimating pose have limited accuracy in different scenes, highlighting the worth of ongoing research on pose estimation from view sequences. Furthermore, our method is based on the assumption of diffuse reflection to learn skin details, which does not perform well in the presence of strong specular reflections. We will continue to explore surface reconstruction for different materials to address this limitation.

\bibliographystyle{IEEEtran}
\bibliography{main}

\begin{thebibliography}{10}
\providecommand{\url}[1]{#1}
\csname url@samestyle\endcsname
\providecommand{\newblock}{\relax}
\providecommand{\bibinfo}[2]{#2}
\providecommand{\BIBentrySTDinterwordspacing}{\spaceskip=0pt\relax}
\providecommand{\BIBentryALTinterwordstretchfactor}{4}
\providecommand{\BIBentryALTinterwordspacing}{\spaceskip=\fontdimen2\font plus
\BIBentryALTinterwordstretchfactor\fontdimen3\font minus \fontdimen4\font\relax}
\providecommand{\BIBforeignlanguage}[2]{{%
\expandafter\ifx\csname l@#1\endcsname\relax
\typeout{** WARNING: IEEEtran.bst: No hyphenation pattern has been}%
\typeout{** loaded for the language `#1'. Using the pattern for}%
\typeout{** the default language instead.}%
\else
\language=\csname l@#1\endcsname
\fi
#2}}
\providecommand{\BIBdecl}{\relax}
\BIBdecl

\bibitem{doosti2020hopenet}
B.~Doosti, S.~Naha, M.~Mirbagheri, and D.~J. Crandall, ``Hope-net: {A} graph-based model for hand-object pose estimation,'' in \emph{IEEE Conf. Comput. Vis. Pattern Recog.}, 2020, pp. 6607--6616.

\bibitem{hasson2020leveraging}
Y.~Hasson, B.~Tekin, F.~Bogo, I.~Laptev, M.~Pollefeys, and C.~Schmid, ``Leveraging photometric consistency over time for sparsely supervised hand-object reconstruction,'' in \emph{IEEE Conf. Comput. Vis. Pattern Recog.}, 2020, pp. 571--580.

\bibitem{fan2021understanding}
H.~Fan, T.~Zhuo, X.~Yu, Y.~Yang, and M.~Kankanhalli, ``Understanding atomic hand-object interaction with human intention,'' \emph{IEEE Trans. Circuit Syst. Video Technol.}, vol.~32, no.~1, pp. 275--285, 2021.

\bibitem{cheng2015survey}
H.~Cheng, L.~Yang, and Z.~Liu, ``Survey on 3d hand gesture recognition,'' \emph{IEEE Trans. Circuit Syst. Video Technol.}, vol.~26, no.~9, pp. 1659--1673, 2015.

\bibitem{SMPL-X}
G.~Pavlakos, V.~Choutas, N.~Ghorbani, T.~Bolkart, A.~A.~A. Osman, D.~Tzionas, and M.~J. Black, ``Expressive body capture: 3d hands, face, and body from a single image,'' in \emph{IEEE Conf. Comput. Vis. Pattern Recog.}, 2019, pp. 10\,975--10\,985.

\bibitem{karunratanakul2022harp}
K.~Karunratanakul, S.~Prokudin, O.~Hilliges, and S.~Tang, ``Harp: Personalized hand reconstruction from a monocular rgb video,'' in \emph{IEEE Conf. Comput. Vis. Pattern Recog.}, 2023, pp. 12\,802--12\,813.

\bibitem{NIMBLE}
Y.~Li, L.~Zhang, Z.~Qiu, Y.~Jiang, N.~Li, Y.~Ma, Y.~Zhang, L.~Xu, and J.~Yu, ``{NIMBLE:} a non-rigid hand model with bones and muscles,'' \emph{ACM Trans. on Graph.}, pp. 120:1--120:16, 2022.

\bibitem{livehand}
A.~Mundra, J.~Wang, M.~Habermann, C.~Theobalt, M.~Elgharib \emph{et~al.}, ``Livehand: Real-time and photorealistic neural hand rendering,'' in \emph{Int. Conf. Comput. Vis.}, 2023, pp. 18\,035--18\,045.

\bibitem{MANO}
J.~Romero, D.~Tzionas, and M.~J. Black, ``Embodied hands: Modeling and capturing hands and bodies together,'' \emph{ACM Trans. on Graph.}, pp. 245:1--245:17, 2017.

\bibitem{SMPL}
M.~Loper, N.~Mahmood, J.~Romero, G.~Pons{-}Moll, and M.~J. Black, ``{SMPL:} a skinned multi-person linear model,'' \emph{ACM Trans. on Graph.}, pp. 248:1--248:16, 2015.

\bibitem{cao2021reconstructing}
Z.~Cao, I.~Radosavovic, A.~Kanazawa, and J.~Malik, ``Reconstructing hand-object interactions in the wild,'' in \emph{Int. Conf. Comput. Vis.}, 2021, pp. 12\,397--12\,406.

\bibitem{MobileHand}
G.~M. Lim, P.~Jatesiktat, and W.~T. Ang, ``Mobilehand: Real-time 3d hand shape and pose estimation from color image,'' in \emph{International Conference on Neural Information Processing}, 2020, pp. 450--459.

\bibitem{alldieck2021imghum}
T.~Alldieck, H.~Xu, and C.~Sminchisescu, ``imghum: Implicit generative models of 3d human shape and articulated pose,'' in \emph{Int. Conf. Comput. Vis.}, 2021, pp. 5441--5450.

\bibitem{ren2024pyramid}
J.~Ren and J.~Zhu, ``Pyramid deep fusion network for two-hand reconstruction from rgb-d images,'' \emph{IEEE Trans. Circuit Syst. Video Technol.}, 2024.

\bibitem{guo20223d}
S.~Guo, E.~Rigall, Y.~Ju, and J.~Dong, ``3d hand pose estimation from monocular rgb with feature interaction module,'' \emph{IEEE Trans. Circuit Syst. Video Technol.}, vol.~32, no.~8, pp. 5293--5306, 2022.

\bibitem{bib:LISA}
E.~Corona, T.~Hodan, M.~Vo, F.~Moreno-Noguer, C.~Sweeney, R.~Newcombe, and L.~Ma, ``Lisa: Learning implicit shape and appearance of hands,'' in \emph{IEEE Conf. Comput. Vis. Pattern Recog.}, 2022, pp. 20\,501--20\,511.

\bibitem{Pose2Mesh}
H.~Choi, G.~Moon, and K.~M. Lee, ``Pose2mesh: Graph convolutional network for 3d human pose and mesh recovery from a 2d human pose,'' in \emph{Eur. Conf. Comput. Vis.}, 2020, pp. 769--787.

\bibitem{chen2021i2uv}
P.~Chen, Y.~Chen, D.~Yang, F.~Wu, Q.~Li, Q.~Xia, and Y.~Tan, ``I2uv-handnet: Image-to-uv prediction network for accurate and high-fidelity 3d hand mesh modeling,'' in \emph{Int. Conf. Comput. Vis.}, 2021, pp. 12\,909--12\,918.

\bibitem{Moon_2020_ECCV_DeepHandMesh}
G.~Moon, T.~Shiratori, and K.~M. Lee, ``Deephandmesh: {A} weakly-supervised deep encoder-decoder framework for high-fidelity hand mesh modeling,'' in \emph{Eur. Conf. Comput. Vis.}, 2020, pp. 440--455.

\bibitem{NeRF}
B.~Mildenhall, P.~P. Srinivasan, M.~Tancik, J.~T. Barron, R.~Ramamoorthi, and R.~Ng, ``Nerf: Representing scenes as neural radiance fields for view synthesis,'' \emph{Communications of the ACM}, pp. 99--106, 2021.

\bibitem{wang2021neus}
P.~Wang, L.~Liu, Y.~Liu, C.~Theobalt, T.~Komura, and W.~Wang, ``Neus: Learning neural implicit surfaces by volume rendering for multi-view reconstruction,'' \emph{Adv. Neural Inform. Process. Syst.}, vol.~34, pp. 27\,171--27\,183, 2021.

\bibitem{bib:HumanNeRF}
C.-Y. Weng, B.~Curless, P.~P. Srinivasan, J.~T. Barron, and I.~Kemelmacher-Shlizerman, ``Humannerf: Free-viewpoint rendering of moving people from monocular video,'' in \emph{IEEE Conf. Comput. Vis. Pattern Recog.}, 2022, pp. 16\,210--16\,220.

\bibitem{SNARF}
X.~Chen, Y.~Zheng, M.~J. Black, O.~Hilliges, and A.~Geiger, ``{SNARF:} differentiable forward skinning for animating non-rigid neural implicit shapes,'' in \emph{Int. Conf. Comput. Vis.}, 2021, pp. 11\,574--11\,584.

\bibitem{liu2021neural}
L.~Liu, M.~Habermann, V.~Rudnev, K.~Sarkar, J.~Gu, and C.~Theobalt, ``Neural actor: Neural free-view synthesis of human actors with pose control,'' \emph{ACM Trans. on Graph.}, pp. 1--16, 2021.

\bibitem{peng2021neural}
S.~Peng, Y.~Zhang, Y.~Xu, Q.~Wang, Q.~Shuai, H.~Bao, and X.~Zhou, ``Neural body: Implicit neural representations with structured latent codes for novel view synthesis of dynamic humans,'' in \emph{IEEE Conf. Comput. Vis. Pattern Recog.}, 2021, pp. 9054--9063.

\bibitem{handnerf}
Z.~Guo, W.~Zhou, M.~Wang, L.~Li, and H.~Li, ``Handnerf: Neural radiance fields for animatable interacting hands,'' in \emph{IEEE Conf. Comput. Vis. Pattern Recog.}, 2023, pp. 21\,078--21\,087.

\bibitem{bib:handavatar}
X.~Chen, B.~Wang, and H.-Y. Shum, ``Hand avatar: Free-pose hand animation and rendering from monocular video,'' in \emph{IEEE Conf. Comput. Vis. Pattern Recog.}, 2023, pp. 8683--8693.

\bibitem{yang2023reconstructing}
G.~Yang, C.~Wang, N.~D. Reddy, and D.~Ramanan, ``Reconstructing animatable categories from videos,'' in \emph{IEEE Conf. Comput. Vis. Pattern Recog.}, 2023, pp. 16\,995--17\,005.

\bibitem{luo2022artemis}
H.~Luo, T.~Xu, Y.~Jiang, C.~Zhou, Q.~Qiu, Y.~Zhang, W.~Yang, L.~Xu, and J.~Yu, ``Artemis: Articulated neural pets with appearance and motion synthesis,'' \emph{ACM Trans. on Graph.}, pp. 164:1--164:19, 2022.

\bibitem{wu2023magicpony}
S.~Wu, R.~Li, T.~Jakab, C.~Rupprecht, and A.~Vedaldi, ``Magicpony: Learning articulated 3d animals in the wild,'' in \emph{IEEE Conf. Comput. Vis. Pattern Recog.}, 2023, pp. 8792--8802.

\bibitem{10.1145/3528223.3530143}
C.~Cao, T.~Simon, J.~K. Kim, G.~Schwartz, M.~Zollh{\"{o}}fer, S.~Saito, S.~Lombardi, S.~Wei, D.~Belko, S.~Yu, Y.~Sheikh, and J.~M. Saragih, ``Authentic volumetric avatars from a phone scan,'' \emph{ACM Trans. on Graph.}, pp. 163:1--163:19, 2022.

\bibitem{zheng2023pointavatar}
Y.~Zheng, W.~Yifan, G.~Wetzstein, M.~J. Black, and O.~Hilliges, ``Pointavatar: Deformable point-based head avatars from videos,'' in \emph{IEEE Conf. Comput. Vis. Pattern Recog.}, 2023, pp. 21\,057--21\,067.

\bibitem{zheng2022avatar}
Y.~Zheng, V.~F. Abrevaya, M.~C. B{\"{u}}hler, X.~Chen, M.~J. Black, and O.~Hilliges, ``I {M} avatar: Implicit morphable head avatars from videos,'' in \emph{IEEE Conf. Comput. Vis. Pattern Recog.}, 2022, pp. 13\,535--13\,545.

\bibitem{grassal2022neural}
P.~Grassal, M.~Prinzler, T.~Leistner, C.~Rother, M.~Nie{\ss}ner, and J.~Thies, ``Neural head avatars from monocular {RGB} videos,'' in \emph{IEEE Conf. Comput. Vis. Pattern Recog.}, 2022, pp. 18\,632--18\,643.

\bibitem{GaoRecon}
X.~Gao, C.~Zhong, J.~Xiang, Y.~Hong, Y.~Guo, and J.~Zhang, ``Reconstructing personalized semantic facial nerf models from monocular video,'' \emph{ACM Trans. on Graph.}, pp. 200:1--200:12, 2022.

\bibitem{yang2022banmo}
G.~Yang, M.~Vo, N.~Neverova, D.~Ramanan, A.~Vedaldi, and H.~Joo, ``Banmo: Building animatable 3d neural models from many casual videos,'' in \emph{IEEE Conf. Comput. Vis. Pattern Recog.}, 2022, pp. 2853--2863.

\bibitem{habermann2021real}
M.~Habermann, L.~Liu, W.~Xu, M.~Zollh{\"{o}}fer, G.~Pons{-}Moll, and C.~Theobalt, ``Real-time deep dynamic characters,'' \emph{ACM Trans. on Graph.}, pp. 94:1--94:16, 2021.

\bibitem{vb-characters}
F.~Xu, Y.~Liu, C.~Stoll, J.~Tompkin, G.~Bharaj, Q.~Dai, H.~Seidel, J.~Kautz, and C.~Theobalt, ``Video-based characters: Creating new human performances from a multi-view video database,'' \emph{ACM Trans. on Graph.}, p.~32, 2011.

\bibitem{peng2022animatable}
S.~Peng, S.~Zhang, Z.~Xu, C.~Geng, B.~Jiang, H.~Bao, and X.~Zhou, ``Animatable neural implicit surfaces for creating avatars from videos,'' \emph{CoRR}, vol. abs/2203.08133, 2022.

\bibitem{bhatnagar2020loopreg}
B.~L. Bhatnagar, C.~Sminchisescu, C.~Theobalt, and G.~Pons-Moll, ``Loopreg: Self-supervised learning of implicit surface correspondences, pose and shape for 3d human mesh registration,'' in \emph{Adv. Neural Inform. Process. Syst.}, 2020, pp. 12\,909--12\,922.

\bibitem{Moon_2020_ECCV_InterHand2.6M}
G.~Moon, S.-I. Yu, H.~Wen, T.~Shiratori, and K.~M. Lee, ``Interhand2.6m: A dataset and baseline for 3d interacting hand pose estimation from a single rgb image,'' in \emph{Eur. Conf. Comput. Vis.}, 2020, pp. 548--564.

\bibitem{pavlakos2023reconstructing}
G.~Pavlakos, D.~Shan, I.~Radosavovic, A.~Kanazawa, D.~Fouhey, and J.~Malik, ``Reconstructing hands in 3d with transformers,'' in \emph{IEEE Conf. Comput. Vis. Pattern Recog.}, 2024, pp. 9826--9836.

\bibitem{boukhayma20193d}
A.~Boukhayma, R.~de~Bem, and P.~H. Torr, ``3d hand shape and pose from images in the wild,'' in \emph{IEEE Conf. Comput. Vis. Pattern Recog.}, 2019, pp. 10\,835--10\,844.

\bibitem{hasson2019learning}
Y.~Hasson, G.~Varol, D.~Tzionas, I.~Kalevatykh, M.~J. Black, I.~Laptev, and C.~Schmid, ``Learning joint reconstruction of hands and manipulated objects,'' in \emph{IEEE Conf. Comput. Vis. Pattern Recog.}, 2019, pp. 11\,807--11\,816.

\bibitem{kong2022identityaware}
D.~Kong, L.~Zhang, L.~Chen, H.~Ma, X.~Yan, S.~Sun, X.~Liu, K.~Han, and X.~Xie, ``Identity-aware hand mesh estimation and personalization from rgb images,'' in \emph{Eur. Conf. Comput. Vis.}, 2022, pp. 536--553.

\bibitem{Ren2023EndtoEndWS}
J.~Ren, J.~Zhu, and J.~Zhang, ``End-to-end weakly-supervised single-stage multiple 3d hand mesh reconstruction from a single rgb image,'' \emph{Computer Vision and Image Understanding}, p. 103706, 2023.

\bibitem{sun2023smr}
H.~Sun, X.~Zheng, P.~Ren, J.~Wang, Q.~Qi, and J.~Liao, ``Smr: Spatial-guided model-based regression for 3d hand pose and mesh reconstruction,'' \emph{IEEE Trans. Circuit Syst. Video Technol.}, vol.~34, no.~1, pp. 299--314, 2023.

\bibitem{li2021latent}
M.~Li, J.~Wang, and N.~Sang, ``Latent distribution-based 3d hand pose estimation from monocular rgb images,'' \emph{IEEE Trans. Circuit Syst. Video Technol.}, vol.~31, no.~12, pp. 4883--4894, 2021.

\bibitem{DBLP:conf/siggraph/OrenN94}
M.~Oren and S.~K. Nayar, ``Generalization of lambert's reflectance model,'' in \emph{Proc. Int. Conf. Comput. Graph. Intera. Tech.}, 1994, pp. 239--246.

\bibitem{chen2022mobrecon}
X.~Chen, Y.~Liu, Y.~Dong, X.~Zhang, C.~Ma, Y.~Xiong, Y.~Zhang, and X.~Guo, ``Mobrecon: Mobile-friendly hand mesh reconstruction from monocular image,'' in \emph{IEEE Conf. Comput. Vis. Pattern Recog.}, 2022, pp. 20\,544--20\,554.

\bibitem{gan2024}
Q.~Gan, W.~Li, J.~Ren, and J.~Zhu, ``Fine-grained multi-view hand reconstruction using inverse rendering,'' in \emph{AAAI}, 2024.

\bibitem{Luan_2023_CVPR}
T.~Luan, Y.~Zhai, J.~Meng, Z.~Li, Z.~Chen, Y.~Xu, and J.~Yuan, ``High fidelity 3d hand shape reconstruction via scalable graph frequency decomposition,'' in \emph{IEEE Conf. Comput. Vis. Pattern Recog.}, 2023, pp. 16\,795--16\,804.

\bibitem{zhu2016video}
H.~Zhu, Y.~Liu, J.~Fan, Q.~Dai, and X.~Cao, ``Video-based outdoor human reconstruction,'' \emph{IEEE Trans. Circuit Syst. Video Technol.}, vol.~27, no.~4, pp. 760--770, 2016.

\bibitem{xavatar}
K.~Shen, C.~Guo, M.~Kaufmann, J.~J. Zarate, J.~Valentin, J.~Song, and O.~Hilliges, ``X-avatar: Expressive human avatars,'' in \emph{IEEE Conf. Comput. Vis. Pattern Recog.}, 2023, pp. 16\,911--16\,921.

\bibitem{sfs}
B.~K.~P. Horn, ``Shape from shading; a method for obtaining the shape of a smooth opaque object from one view,'' Ph.D. dissertation, Massachusetts Institute of Technology, {USA}, 1970.

\bibitem{Laine2020diffrast}
S.~Laine, J.~Hellsten, T.~Karras, Y.~Seol, J.~Lehtinen, and T.~Aila, ``Modular primitives for high-performance differentiable rendering,'' \emph{ACM Trans. on Graph.}, pp. 194:1--194:14, 2020.

\bibitem{aliev2020neural}
K.~Aliev, A.~Sevastopolsky, M.~Kolos, D.~Ulyanov, and V.~S. Lempitsky, ``Neural point-based graphics,'' in \emph{Eur. Conf. Comput. Vis.}, 2020, pp. 696--712.

\bibitem{bib:nccsfs}
L.~Lin, S.~Peng, Q.~Gan, and J.~Zhu, ``Fasthuman: Reconstructing high-quality clothed human in minutes,'' in \emph{International Conference on 3D Vision}, 2024.

\bibitem{nealen2006laplacian}
A.~Nealen, T.~Igarashi, O.~Sorkine, and M.~Alexa, ``Laplacian mesh optimization,'' in \emph{Proc. Int. Conf. Comput. Graph. Intera. Tech.}, 2006, pp. 381--389.

\bibitem{kerbl20233d}
B.~Kerbl, G.~Kopanas, T.~Leimk{\"u}hler, and G.~Drettakis, ``3d gaussian splatting for real-time radiance field rendering,'' \emph{ACM Trans. on Graph.}, pp. 1--14, 2023.

\bibitem{chan2022efficient}
E.~R. Chan, C.~Z. Lin, M.~A. Chan, K.~Nagano, B.~Pan, S.~De~Mello, O.~Gallo, L.~J. Guibas, J.~Tremblay, S.~Khamis \emph{et~al.}, ``Efficient geometry-aware 3d generative adversarial networks,'' in \emph{IEEE Conf. Comput. Vis. Pattern Recog.}, 2022, pp. 16\,123--16\,133.

\bibitem{unet}
O.~Ronneberger, P.~Fischer, and T.~Brox, ``U-net: Convolutional networks for biomedical image segmentation,'' in \emph{Medical Image Computing and Computer-Assisted Intervention}, 2015, pp. 234--241.

\end{thebibliography}
% \end{thebibliography}

\begin{IEEEbiography}[{\includegraphics[width=1in,height=1.25in,clip,keepaspectratio]{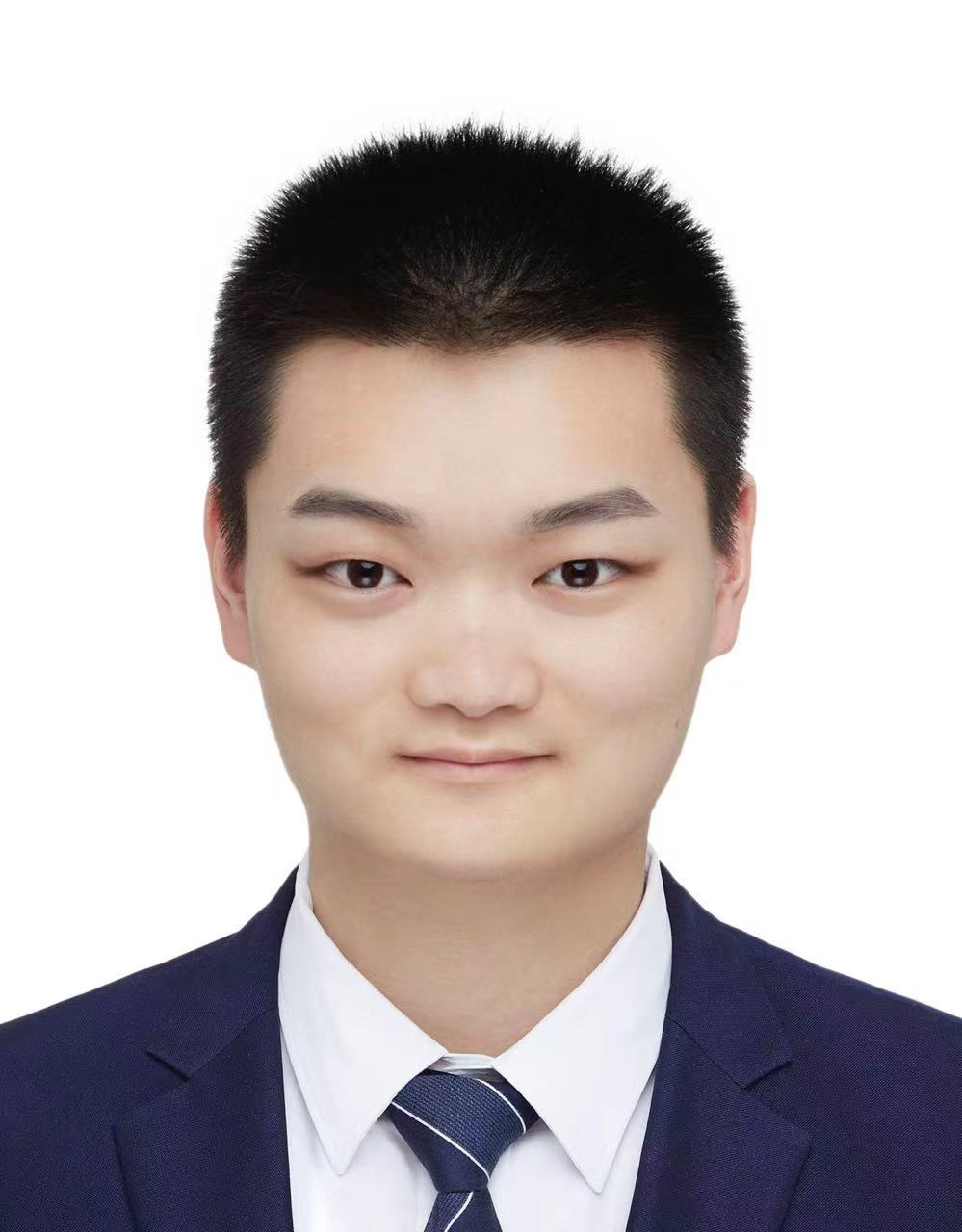}}]{Qijun Gan} is currently a PhD candidate in the College of Computer Science and Technology, Zhejiang University, Hangzhou, China. Before that, he received the bachelor degree from University of International Business and Economics, China. His research interests include machine learning and computer vision, with a focus on 3D reconstruction.\end{IEEEbiography}

\begin{IEEEbiography}[{\includegraphics[width=1in,height=1.25in,clip,keepaspectratio]{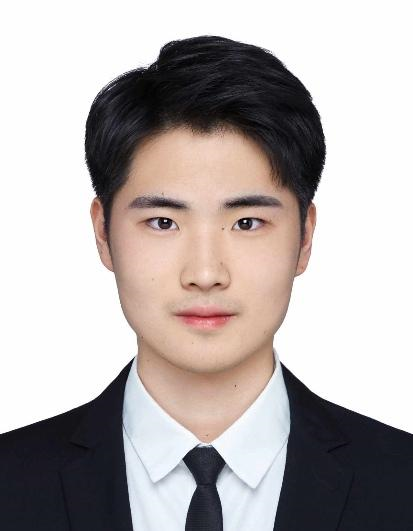}}]{Zijie Zhou} received the B.S. degree in Communication Engineering from Beijing University of Post and Telecommunication, Beijing, China, in 2022. He is currently a postgraduate student in the School of Software Technology, Zhejiang University, Hangzhou, China. His research interests include computer vision and deep learning.
\end{IEEEbiography}

\begin{IEEEbiography}[{\includegraphics[width=1in,height=1.25in,clip,keepaspectratio]{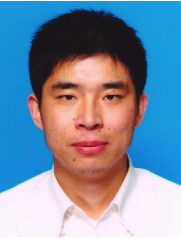}}]{Jianke Zhu}{\,} received the master’s degree from University of Macau in Electrical and Electronics Engineering, and the PhD degree in computer science and engineering from The Chinese University of Hong Kong, Hong Kong in 2008. He held a post-doctoral position at the BIWI Computer Vision Laboratory, ETH Zurich, Switzerland. He is currently a Professor with the College of Computer Science, Zhejiang University, Hangzhou, China. His research interests include computer vision and robotics. 
\end{IEEEbiography}

\end{document}